\documentclass[journal]{IEEEtran}
\usepackage{amsmath,amsfonts}
\usepackage{algorithmic}
\usepackage{array}
\usepackage[justification=justified,singlelinecheck=true]{caption}
\usepackage{textcomp}
\usepackage{stfloats}
\usepackage{url}
\usepackage{verbatim}
\usepackage{graphicx}
\usepackage{subcaption}
\usepackage{cite}
\usepackage{booktabs}       
\usepackage{multirow}
\usepackage{tabularx}
\usepackage{graphicx}       
\usepackage{colortbl}       
\usepackage{hyperref}
\usepackage{kotex}          
\hypersetup{
    colorlinks=true,
    urlcolor=red,
}
\usepackage[final,nopatch=footnote]{microtype}

\usepackage{graphicx}
\usepackage{float}
\usepackage{lscape}                                         
\usepackage{wrapfig}
\usepackage[lined,ruled,linesnumbered]{algorithm2e}
\usepackage{animate} 

\usepackage{booktabs}                   
\usepackage{multirow}

\usepackage{makecell}

\usepackage{paralist}
\usepackage{enumitem}

\usepackage{bm}                          
\usepackage{epsfig}                      
\usepackage{graphicx}                  
\usepackage{times}
\usepackage{mathptmx}
\usepackage{mathtools}
\usepackage{amssymb,amsmath}   

\usepackage{units}
\usepackage{color}
\usepackage[T1]{fontenc}    
\usepackage{amsfonts}       
\usepackage[utf8]{inputenc} 
\usepackage{nicefrac}       
\usepackage{microtype}      
\usepackage{comment}

\usepackage{url}  
\usepackage{xspace}
\usepackage[table]{xcolor}
\usepackage{setspace}
\usepackage{grfext}
\PrependGraphicsExtensions*{.jpg,.png,.PNG}

\usepackage{amssymb}
\usepackage{pifont}

\def\eg{e.g.,~}               
\def\ie{i.e.,~}               
\def\vs{vs.~}                 

\newlength\paramargin
\newlength\figmargin

\newlength\secmargin
\newlength\figcapmargin
\newlength\tabcapmargin

\newcommand{\alwaysred}{\textcolor{red}}
\newcommand{\red}{}
\newcommand{\sred}{} 

\newcommand{\figcaption}[2]
{
\caption{
\textbf{#1.}  
#2            
}
}

\newcommand{\secref}[1]{Section~\ref{sec:#1}}
\newcommand{\figref}[1]{Figure~\ref{fig:#1}} 
\newcommand{\tabref}[1]{Table~\ref{tab:#1}}

\newcommand{\eqnref}[1]{\eqref{eq:#1}}

\long\def\ignorethis#1{}

\newcommand{\caast}{C$\text{A}^2$ST}

\newcommand{\tb}[1]{\textbf{#1}}

\newbox\jsavebox%

\newcommand{\best}[1]{{\textbf{#1}}}
\newcommand{\second}[1]{{\underline{#1}}}

\def\xi{\mathbf{x}_i}

\def\I{\mathbf{I}}
\def\B{\mathbf{B}}
\def\E{\mathbf{E}}
\def\X{\mathbf{X}}
\def\Y{\mathbf{Y}}
\def\Z{\mathbf{Z}}
\def\W{\mathbf{W}}

\begin{document}

\title{\texorpdfstring{CA$^2$ST}{CA2ST}: Cross-Attention in Audio, Space, and Time for Holistic Video Recognition}
\author{
    \IEEEauthorblockN{Jongseo Lee\IEEEauthorrefmark{1}, Joohyun Chang\IEEEauthorrefmark{1}, Dongho Lee, Jinwoo Choi\IEEEauthorrefmark{2}}
    \\
    \IEEEauthorblockA{Kyung Hee University, Republic of Korea
    \\\{jong980812, joohyun7u, kide004, jinwoochoi\}@khu.ac.kr}
\thanks{This paper was produced by the IEEE Publication Technology Group. They are in Piscataway, NJ.}
\thanks{Manuscript received September 24, 2024; revised September 30, 2025; accepted October 31, 2025.}
\thanks{\IEEEauthorrefmark{1}Equally contributed first authors.}
\thanks{\IEEEauthorrefmark{2}Corresponding author.}
}
\markboth{IEEE Transactions on Pattern Analysis and Machine Intelligence}%
{Shell \MakeLowercase{\textit{et al.}}: A Sample Article Using IEEEtran.cls for IEEE Journals}


\maketitle

\begin{abstract}

We propose Cross-Attention in Audio, Space, and Time (C$\text{A}^2$ST), a transformer-based method for holistic video recognition.
Recognizing actions in videos requires both spatial and temporal understanding, yet most existing models lack a balanced spatio-temporal understanding of videos.
To address this, we propose a novel two-stream architecture, called Cross-Attention in Space and Time (CAST), using only RGB input.
In each layer of CAST, Bottleneck Cross-Attention (B-CA) enables spatial and temporal experts to exchange information and make synergistic predictions.
For holistic video understanding, we extend CAST by integrating an audio expert, forming Cross-Attention in Visual and
Audio (CAVA). 
We validate the CAST on benchmarks with different characteristics, EPIC-KITCHENS-100, Something-Something-V2, Kinetics-400, \red{
ActivityNet, and HD-EPIC to show balanced performance. 
}
We also validate the CAVA on audio-visual action recognition benchmarks, including UCF-101, VGG-Sound, KineticsSound, EPIC-
SOUNDS, \red{
and HD-EPIC-SOUNDS.
}
CAVA shows favorable performance on these datasets, demonstrating the effective information exchange among multiple experts within the B-CA module.
In addition, C$\text{A}^2$ST combines CAST and CAVA by employing spatial, temporal, and audio experts through cross-attention, achieving balanced and holistic video understanding.
\end{abstract}


\vspace{-0.2em}
\section{Introduction}
\label{sec:intro}
\begin{figure*}[!t]
\centering
\includegraphics[width=0.9\textwidth]{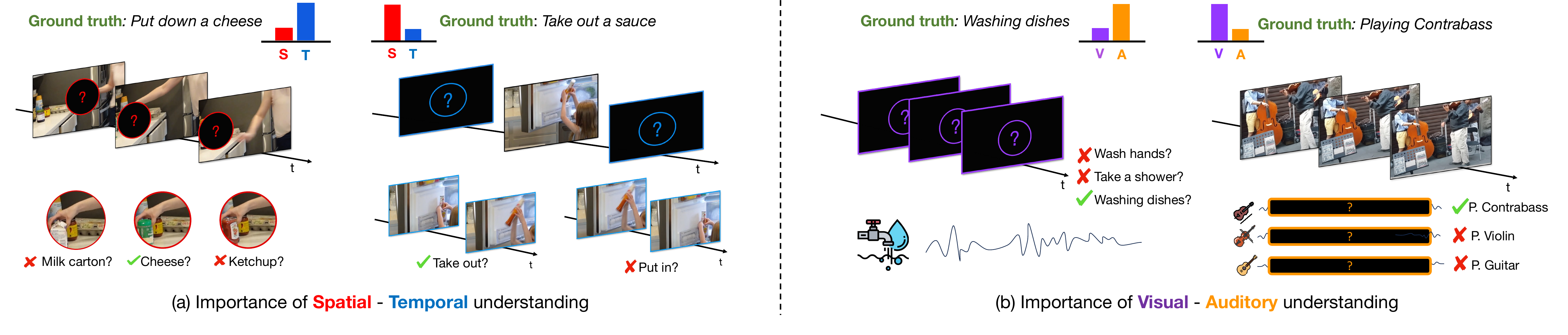}

\vspace{-0.5em}
\figcaption{Importance of balanced spatio-temporal and audio-visual understanding}
{
(a) A model without fine-grained spatial understanding fails to predict \emph{Put down cheese}, due to subtle appearance differences between the objects. 
On the other hand, if a model lacks sufficient temporal context understanding, the model may incorrectly predict an action. 
For example, the model without temporal context understanding fails to predict \emph{Take out sauce} since it is hard to distinguish from \emph{Put in} without knowing order of the events. 
(b) If a model lacks audio cues, it may struggle to distinguish between different musical instruments: \emph{contrabass} versus \emph{violin} because of their visual similarities. 
On the other hand, when visual cues are missing and only audio cues are available, distinguishing actions such as \emph{wash hands} and \emph{washing dishes} becomes challenging due to similar sounds. 
}
\label{fig:teaser}
\vspace{-1.5em}
\end{figure*}

\IEEEPARstart{V}{ideo} recognition involves two key contexts: spatial and temporal, which together form the spatio-temporal context.
To accurately recognize human actions in videos, a model must understand both the spatial and temporal contexts. 
In \figref{teaser} (a), we show the importance of understanding both contexts together. 
A model without fine-grained spatial understanding is likely to fail, such as confusing whether an object in the hand is a \emph{milk carton}, or \emph{cheese}, or \emph{ketchup}. 
Similarly, a model without temporal context understanding may fail to recognize the sequence of actions, for instance, whether the hand is moving into or out of a fridge. 
Therefore, for accurate action recognition, a model needs to balance its understanding of both spatial and temporal contexts.

Despite the recent progress in action recognition through the use of Transformers~\cite{vaswani2017attention,dosovitskiy2020vit,bertasius2021space,tong2022videomae}, achieving a balanced spatio-temporal understanding remains a challenging problem. 
Due to the lack of sufficient video training data~\cite{bertasius2021space}, many video models~\cite{yang2023aim,pan2022st,lin2022frozen} extended from the image domain rely heavily on pre-trained image models, leading to inadequate temporal modeling. 
On the other hand, models trained from scratch on video datasets tend to relatively lack a spatial understanding~\cite{tong2022videomae,wang2022bevt} due to additional temporal dimension during pre-training stage.    
Consequently, most action recognition models lack a balanced spatio-temporal understanding of videos. 

To further investigate the imbalanced performances, we analyze how models perform across diverse datasets with different characteristics.
We find that models that perform well on static-biased~\cite{li2018resound,Choi-NeurIPS-2019,sevilla2021only} datasets, such as Kinetics-400, may not perform as well on temporal-biased~\cite{bertasius2021space,kowal2022deeper} datasets, such as Something-Something-V2, and vice versa.
For instance, as shown in \tabref{main_rgb}, BEVT~\cite{wang2022bevt} outperforms AIM~\cite{yang2023aim} on the Something-Something-V2 dataset, while BEVT underperforms AIM on the Kinetics-400 dataset. We observe a similar trend between ST-Adapter~\cite{pan2022st} and VideoMAE~\cite{tong2022videomae} on the EPIC-KITCHENS-100~\cite{damen2022rescaling} dataset, where one model performs better on noun prediction while the other excels in verb prediction.

One possible solution to the challenge of balanced spatio-temporal understanding is to use multi-modal learning. For example, two-stream networks~\cite{Simonyan-NIPS-2014,feichtenhofer2016convolutional} employ both RGB and optical flow streams to learn both spatial and temporal contexts. However, this approach can be computationally expensive due to optical flow estimation. 

In this work, we introduce a two-stream architecture, Cross-Attention in Space and Time (CAST), to address the challenge of balanced spatio-temporal understanding using only RGB input.  
CAST employs two expert models - a spatial expert and a temporal expert - which exchange information to make a synergistic collective prediction. 
More specifically, we realize the information exchange by cross-attention between the two experts in the Bottleneck Cross-Attention (B-CA) module. 
As a result, CAST effectively achieves synergy through the information exchange between spatial and temporal experts.

To achieve a more holistic video understanding capability, we extend CAST to incorporate audio information as well as visual information.
In many cases, audio information could provide valuable context that visual information alone may not capture~\cite{keysers2003audiovisual}. 
In \figref{teaser} (b), we highlight the importance of understanding both visual and audio information for holistic video recognition.
For example, distinguishing between different musical instruments, such as a \emph{contrabass} versus a \emph{violin}, could be challenging if a model solely relies on visual information due to the visual similarity between the two instruments. 
Since the sound of the two instruments is quite distinctive, using audio information could be beneficial for recognizing a video more holistically.
Conversely, when visual cues are missing but the audio is available, it is challenging to distinguish between \emph{Wash Hands} and \emph{Washing Dishes} actions by relying solely on the sound of running water. 
The absence of visual information makes it difficult to accurately identify the action. 
Therefore, for holistic video action recognition, it is crucial to consider both visual and audio modalities.

In this work, we extend CAST~\cite{lee2024cast}, our previous work, to incorporate audio information as well.
Since the proposed B-CA module facilitates information exchange between experts in different contexts \ie{space and time}, it is natural to apply the B-CA module for cross-modal information exchange between experts from visual and audio modalities.
As a result, we introduce Cross-Attention in Visual and Audio (CAVA), which leverages cross-attention mechanisms to facilitate synergistic learning between visual and audio expert models.
%

\begin{figure}[!t]
\centering
\includegraphics[width=0.9\columnwidth]{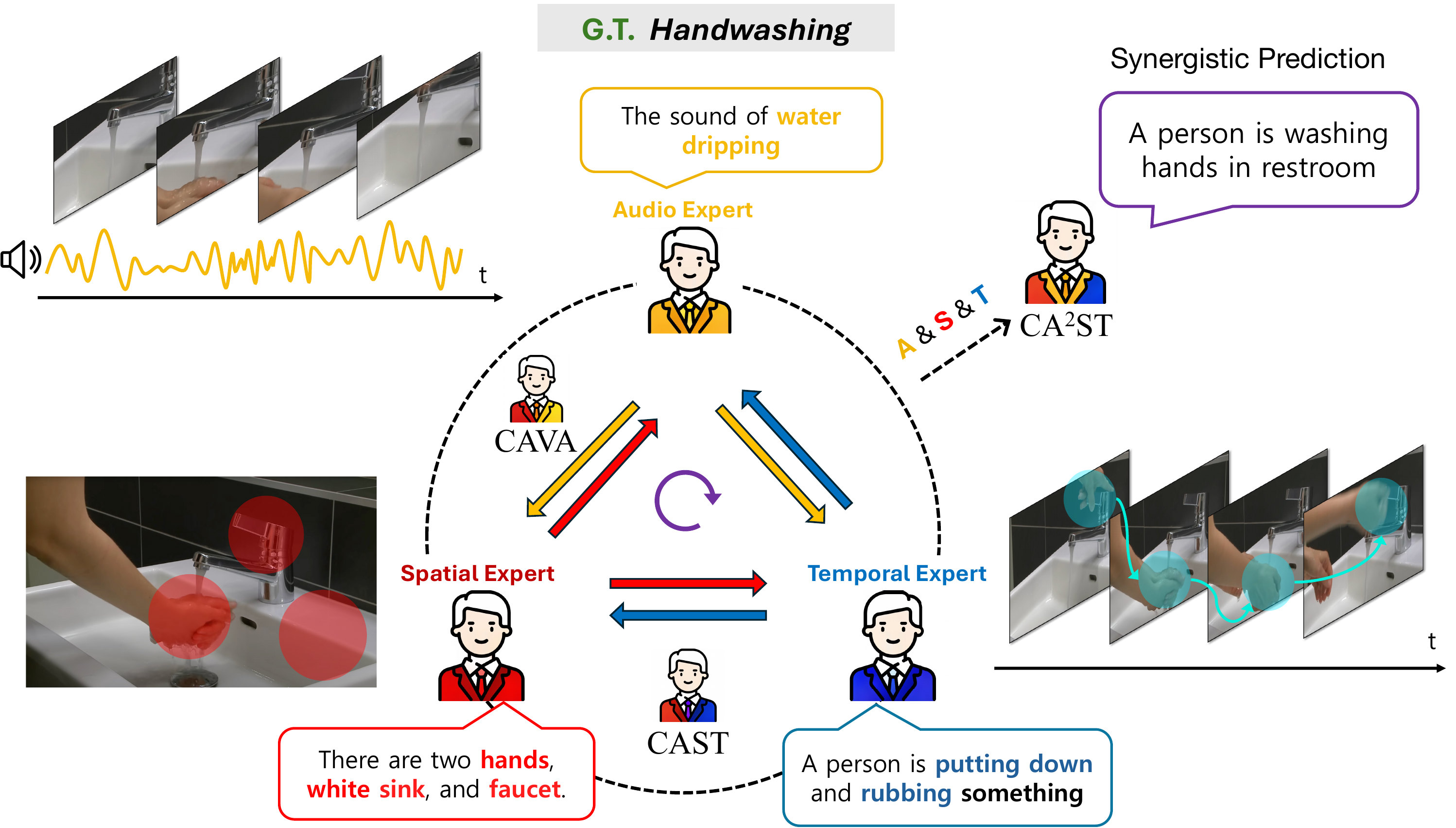}

\figcaption{High-level illustration of CA$^2$ST}{
C$\text{A}^2$ST employs spatial, temporal, and audio expert models.
The three experts exchange information through interactions and they teach each other. 
In the early stage, the experts might be able predict only partial information due to the lack of comprehensive understanding. 
After multiple iterations of information exchange among the experts, the proposed method can collectively predict the correct action: \emph{Washing hands in the restroom}.
}

\vspace{-1.5em}
\label{fig:motivation}
\end{figure}

In \figref{motivation}, we present a high-level illustration of our proposed methods: CAVA and C$\text{A}^2$ST.
CAVA employs a visual (spatial) expert and an audio expert and achieves synergistic multimodal understanding through information exchange between them.
Furthermore, by employing both spatial and temporal experts for the visual expert, all three experts \ie{space, time, and audio} can exchange information with each other. 
We refer to this configuration as C$\text{A}^2$ST (Cross-Attention in Audio, Space, and Time), where cross-attention actively facilitates the exchange of information among the three experts.
CAVA and C$\text{A}^2$ST integrate audio information essential for holistic video recognition by facilitating information exchange with the audio expert.
Our method can effectively learn from multiple modalities without the extra need for pre-training on large multi-modal datasets. 
By leveraging both visual and audio data, we improve the overall video understanding performance.

To validate the effectiveness of the CAST, we conduct extensive experiments on multiple action recognition datasets with distinct characteristics, including the temporal-biased Something-Something-V2~\cite{goyal2017something}, static-biased Kinetics-400~\cite{kay2017kinetics}, fine-grained EPIC-KITCHENS-100~\cite{damen2022rescaling}, \red{ 
ActivityNet~\cite{caba2015activitynet} with longer videos, and 
HD-EPIC~\cite{hd-epic}.
}
Our results demonstrate that CAST achieves balanced spatio-temporal understanding and shows favorable performance across these different datasets.
We also validate CAVA and C$\text{A}^2$ST on four audio-visual datasets, including UCF-101~\cite{soomro2012dataset}, VGG-Sound~\cite{chen2020vggsound}, KineticsSound~\cite{xiao2020audiovisual}, 
\red{EPIC-SOUNDS~\cite{huh2023epic}, and HD-EPIC-SOUNDS. 
}
Our experiments demonstrate that CAVA and C$\text{A}^2$ST effectively leverage both audio and visual modality, outperforming the models using only each modality and other multi-modal methods.


In this work, we make the following significant contributions.
\begin{itemize}
    \item We introduce a two-stream architecture, CAST, which addresses the challenge of \emph{balanced spatio-temporal understanding} that has been largely overlooked by previous works.
    \item We conduct extensive experiments on multiple datasets with distinct characteristics to demonstrate the effectiveness of CAST. In terms of \emph{balanced} spatio-temporal understanding, CAST shows favorable performance, while existing methods show more imbalanced performance.
    \item We conduct an extensive ablation study and analysis to validate the design choices of the proposed method. We show that employing \emph{spatial expert} and \emph{temporal expert} and applying \emph{cross-attention in a bottleneck} architecture between the experts is crucial for achieving effective spatio-temporal representation learning.
    \item We extend CAST into a multi-modal framework, CAVA, which integrates both visual and audio experts to further enhance action recognition capabilities. Additionally, we introduce C$\text{A}^2$ST, which utilizes three experts \eg{audio, spatial, and temporal} allowing for comprehensive information exchange to achieve holistic video understanding. 
    \item We validate CAVA and C$\text{A}^2$ST on multiple audio-visual datasets, demonstrating the ability to effectively leverage audio cues, thereby enhancing the model’s robustness and enabling more holistic video action recognition.
\end{itemize}

\vspace{\secmargin}
\section{Related Work}
\label{sec:related}

\subsection{Video action recognition.} 
CNN-based approaches have been extensively applied to action recognition, including 2D CNNs~\cite{wang2018temporal,zhou2018temporal,lin2019tsm}, 3D CNNs\sred{~\cite{tran2015learning, carreira2017quo,tran2018closer,tu2019action,yang2024cross}}, 2D and 1D separable CNNs~\cite{tran2018closer, Xie-ECCV-2018}, or two-stream CNNs~\cite{feichtenhofer2016convolutional,feichtenhofer2019slowfast}. 
With the strong inductive biases, these methods have achieved great progress. 
Recently, Transformer-based approaches~\cite{arnab2021vivit, bertasius2021space, herzig2022object, patrick2021keeping, wu2022memvit, fan2021multiscale, yan2022multiview} become popular in the community due to the long-term context modeling capabilities.
Similar to the two-stream CNNs, we propose a two-stream Transformer architecture, CAST, consisting of two expert models. Unlike conventional two-stream CNNs, CAST uses solely RGB input, without incorporating both RGB and optical flow.

\vspace{\paramargin}
\subsection{Cross-modal fusion}
\sred{Recent advances in computational resources and the emergence of large-scale, high-quality multimodal datasets have significantly facilitated research on cross-modal fusion.
As a result, various approaches~\cite{liu2024visual,li2023logonet,zhao2023learning,joint-bone,lavish, yang2024cross,chen2024internvl,wang2024internvideo2,alayrac2022flamingo,li2023blip,tu2019action} have been proposed to effectively align different modalities such as RGB, text, and skeleton, demonstrating strong performance across diverse benchmarks.}

Among these approaches for cross-modal fusion, cross-attention has emerged as a particularly effective mechanism for facilitating information exchange across modalities, including audio, vision, and text~\cite{nagrani2021mbt,zhao2023learning,alayrac2022flamingo,li2023blip,han2024noise}.
Recently, cross-attention between different views of the same video has shown impressive results~\cite{yan2022multiview,Zhu_2022_CVPR,Kim_2022_ACCV}.
Similar to these, we propose the B-CA module, which utilizes a cross-attention mechanism within a bottleneck architecture. The B-CA module facilitates effective information exchange between two distinct expert models. 
The proposed framework, C$\text{A}^2$ST uses the B-CA module for bidirectional cross-attention, to achieve a balanced spatio-temporal and audio-visual understanding.

\vspace{\paramargin}
\subsection{Foundation model.}
Trained on web-scale datasets using self-supervised learning, foundation models~\cite{brown2020language,radford2021learning,chu2024visionllama,wang2024internvideo2,siglip} are highly adaptable and versatile. 
By leveraging multi-modal learning, foundation models demonstrate strong performance across various tasks, showing impressive results in visual, textual, and audio domains, including computer vision~\cite{wang2022internvideo,wang2022bevt}, natural language processing~\cite{scao2022bloom,touvron2023llama}, and audio recognition~\cite{girdhar2023imagebind,liu2024audioldm}.
In this work, as one realization of C$\text{A}^2$ST, we employ CLIP~\cite{radford2021learning} as our spatial expert because it shows remarkable performance on more than 30 computer vision tasks.

Recent trends in foundation models have shifted from encoder-only architectures~\cite{tong2022videomae,XCLIP,yang2023aim,wang2022internvideo} 
to multimodal systems that integrate large language models (LLMs) with visual or audio encoders, 
often referred to as video large language models (VidLLMs)~\cite{videollm,video-chatgpt,video-llava,chu2024visionllama,wang2024internvideo2}. 
In this work, we position CAST/CAVA/CA$^2$ST as efficient video encoders with bottleneck cross-attention, 
complementary to VidLLMs and readily applicable as drop-in backbones for spatio-temporal and audio-visual representation learning.

\vspace{\paramargin}
\subsection{Parameter-efficient transfer learning.}
\emph{Pre-training then fine-tuning} approaches using powerful foundation models have achieved impressive results in various computer vision tasks. 
However, fully fine-tuning the entire model is often computationally expensive and unnecessary~\cite{yang2023aim}.
Updating only a small subset of parameters while keeping the rest frozen can be effective for both NLP~\cite{karimi2021compacter,lester2021power} and computer vision tasks~\cite{wang2022learning,rebuffi2017learning}.
In particular, extending image foundation models with adapter architectures is effective in action recognition~\cite{lin2022frozen,yang2023aim,pan2022st}. 
We also incorporate an adapter architecture with cross-attention between two expert models. 
We empirically demonstrate that the proposed method outperforms existing adapter-based video models in achieving balanced spatio-temporal and audio-visual understanding.

\vspace{\paramargin}
\subsection{Audio-visual action recognition. }
Many works\sred{~\cite{xiao2020audiovisual,nagrani2021mbt,huang2024mavil,chalk2024tim,han2024noise}} have addressed the challenge of audio-visual understanding in videos. 
These works show the importance of both audio and visual information to enhance action recognition performance.   
Additionally, many researchers have directly pre-trained large models~\cite{piergiovanni2024mirasol3b,lu2024unified} on extensive datasets to achieve comprehensive audio-visual understanding. 
However, these approaches require significant computation costs due to the need for extensive pre-training on large-scale datasets.
In contrast, our approach efficiently leverages already pre-trained models through parameter-efficient tuning, thus achieving effective audio-visual action recognition. 
In this work, we use AST~\cite{gong21b_interspeech} as our audio expert, which is a transformer-based architecture pre-trained on AudioSet~\cite{gemmeke2017audio} using an audio spectrogram as an input.

\section{Cross-Attention in Audio, Space, and Time}
\label{sec:method}
\begin{figure*}[!t]
\centering
\includegraphics[width=0.8\textwidth]{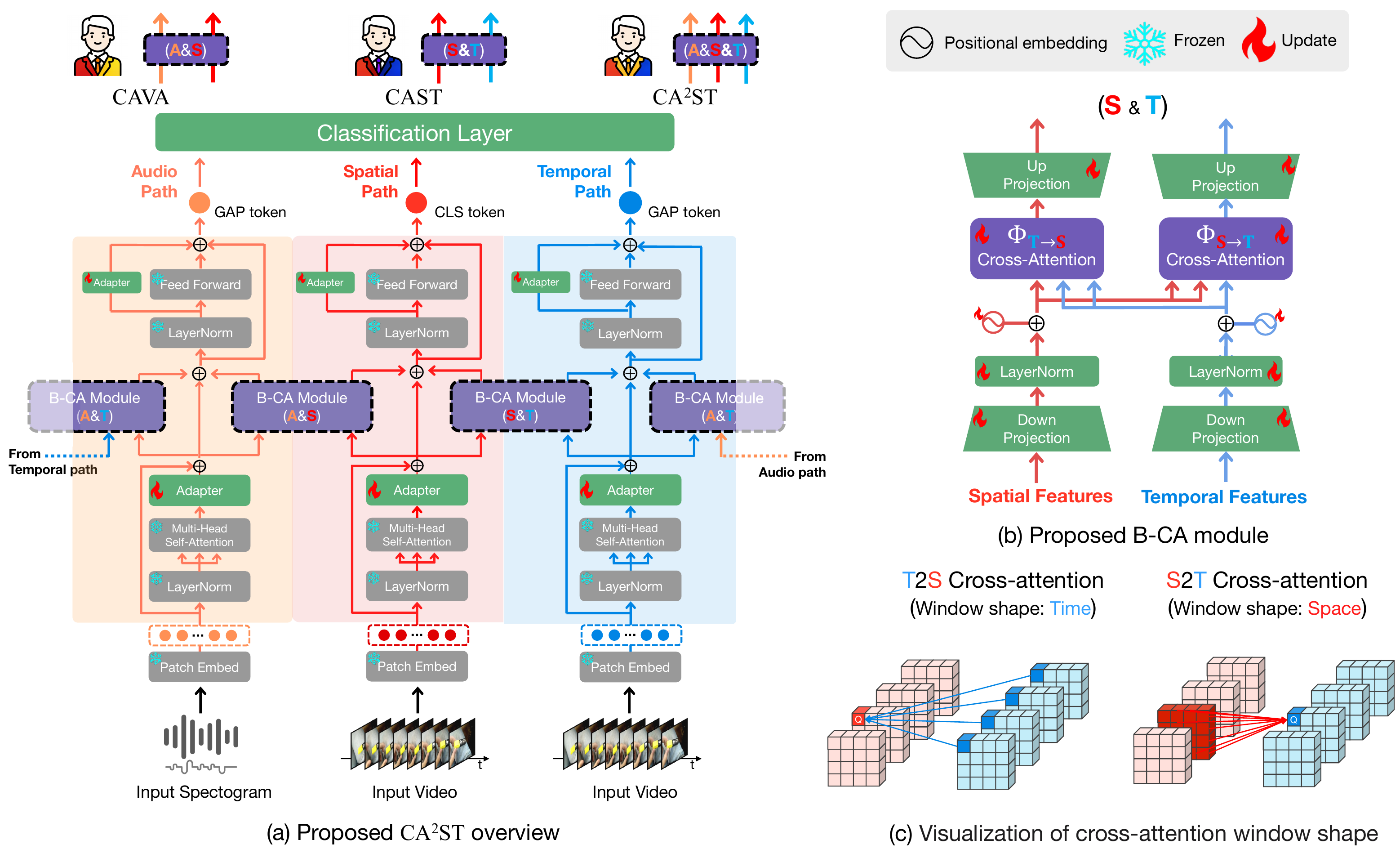}

\figcaption{Overview of CA$^2$ST}{
(a) CAST employs frozen spatial and temporal experts, connected through bottleneck cross-attention (B-CA) modules that facilitate information exchange between the two paths (\textbf{S\&T}).
CAVA employs frozen audio and visual (spatial) experts and the two experts exchange information via B-CA modules (\textbf{A\&S}).
\caast{} extends this architecture by incorporating three paths (\emph{spatial}, \emph{temporal}, and \emph{audio}) connected through three B-CA modules (\textbf{A\&S}, \textbf{S\&T}, \textbf{A\&T}). 
For better adaptation, we learn only the small number of parameters from the B-CA modules and adapters while we freeze all the other parameters.
(b) For simplicity, we illustrate only the \textbf{S\&T} B-CA module, as the other types of B-CA module differ only in the experts. 
The \textbf{S\&T} B-CA module enables temporal-to-spatial (T2S) and spatial-to-temporal (S2T) cross-attentions, facilitating a balanced understanding of spatio-temporal features. 
To enable efficient and effective learning, we incorporate cross-attention into the bottleneck adapter. 
We employ separate position embedding for each expert.
(c) 
In T2S, the model attends along the temporal axis only. In contrast, in S2T, the model attends along the spatial axes only. }
\vspace{\figcapmargin}
\label{fig:overview}
\end{figure*}

C$\text{A}^2$ST is a multi-modal learning architecture consisting of audio, space, and time expert models as illustrated in \figref{overview}.
We first introduce CAST~\cite{lee2024cast}, a balanced spatio-temporal representation learning method for action recognition.
As shown in \figref{overview} (a), excluding the audio path, CAST leverages frozen spatial and temporal expert models.
We can select any vision transformer, each consisting of 12 transformer blocks, as a spatial/temporal expert model. 
To facilitate information exchange between the experts, we incorporate bottleneck cross-attention (B-CA) modules on top of the frozen layers of each expert model.
Through the B-CA modules, the experts exchange information bidirectionally, leading to a more balanced spatio-temporal understanding capability compared to employing independent experts. 
Similar to a prior work~\cite{yang2023aim}, CAST also employs adapter layers with a small number of learnable parameters to improve adaptation to downstream tasks. 

To achieve holistic video understanding capabilities, we extend CAST to incorporate audio modality as well.
To this end, we introduce CAVA with the \emph{audio path} as shown in \figref{overview} (a).
CAVA consists of one visual (spatial) expert model and one audio expert model and they exchange information through the B-CA modules. 
As depicted in \figref{overview} (a), C$\text{A}^2$ST combines CAST and CAVA, using audio, spatial, and temporal experts.
C$\text{A}^2$ST creates synergy through cross-attention among all three expert models.
As a result, C$\text{A}^2$ST achieves a more balanced and holistic video understanding capability.
In the following subsections, we provide a detailed description of each component of our proposed architectures.

\vspace{-0.5em}
\vspace{\paramargin}
\subsection{Input embeddings}
\label{sec:embedding}
\subsubsection{RGB input}
An input to CAST is a mini-batch $\I_{\text{video}} \in \mathbb{R}^{B\times 2T\times H\times W\times C}$, consisting of $B$ RGB videos of $2T$ frames, $H \times W$ spatial dimension, and $C$ channels.
We apply patch tokenization to the input videos for the spatial expert and the temporal expert.
For the spatial expert, we split every even frame of each video in a minibatch into $N$ non-overlapping patches of $p \times p$ pixels~\cite{dosovitskiy2020vit}. 
Then we feed the patches into a frozen linear layer and add position embeddings to obtain spatial embeddings, $\X_s \in \mathbb{R}^{BT \times N \times D}$, where $D$ represents the patch dimension. 
For the temporal expert, we split every two frames of each video in a minibatch into $2 \times p \times p$ pixels non-overlapping tubes~\cite{arnab2021vivit}. 
Then we feed the tubes into a frozen linear layer and add position embeddings to obtain temporal embeddings, $\X_t \in \mathbb{R}^{B \times TN \times D}$.

\subsubsection{Audio input}
To feed as an audio input to CAVA, we first convert an audio waveform of $t$ seconds into a $128$-dimensional Mel spectrogram.
CAVA takes RGB videos and the corresponding time-aligned spectrograms as inputs.
We construct a minibatch of spectrograms, $\I_{\text{audio}} \in \mathbb{R}^{B \times T \times \Omega}$, where $B$ is the batch size, $T$ and $\Omega$ are the time and frequency dimensions of the spectrogram, respectively. 
Following a prior work~\cite{gong21b_interspeech}, we apply patch tokenization by dividing the spectrogram into $M$ overlapping patches of $16 \times 16$ pixels, with an overlap of 6 pixels in both time and frequency dimensions.
Then we feed the patches into a frozen linear layer and add positional embeddings to obtain audio embeddings, $\X_a \in \mathbb{R}^{B \times M \times D}$, where $D$ represents the patch dimension.
\vspace{\paramargin}
\subsection{CAST architecture} 
\vspace{-0.2em}
\label{sec:architecture_cast}
For simplicity, we first describe the CAST architecture in detail, including the key operations of C$\text{A}^2$ST. 
We describe the extensions of CAST architecture to CAVA and \caast{} in detail in \secref{architecture_cava} and \secref{caast}, respectively.
The model architecture of each expert is the same as the ViT~\cite{dosovitskiy2020vit} except for adapters and the B-CA module.
Only the parameters of adapters, B-CA, and classification layer are learnable while we freeze all the other parameters. 

\subsubsection{Original ViT layer}
For completeness, we first define the operations used and then describe the entire model architecture.
Given an input $\X$, we define Multi-Head Self Attention (MHSA) operation as follows:
\begin{equation}
    \label{eq:mhsa}
    \text{MHSA}(\X)=\text{Softmax}((\X \W_{Q}) (\X \W_{K})^{\top}) (\X \W_{V}),
\end{equation}
\noindent
where $\text{Softmax}(\cdot)$ denotes the softmax function, and $\W_{Q}$, $\W_{K}$, and $\W_{V}$ are the query, key, and value projection matrices, respectively.
We also define the adapter operation with linear down and up projection matrices $\W_{D}$ and $\W_{U}$ as follows:
\begin{equation}
\label{eq:adap}
\text{ADAP}(\X) = \sigma(\X \W_{D}) \W_{U},
\end{equation}
\noindent
where $\sigma(\cdot)$ is the GELU activation function~\cite{hendrycks2016gaussian}.

For each attention block $l$, we apply independent Multi-Head Self Attention (MHSA) for each expert along with a skip connection as follows:
\begin{align}
\Y^{(l)} &= \X^{(l)} + \text{ADAP} (\text{MHSA}(\text{LN}(\X^{( l)}))) + \text{MHSA}(\text{LN}(\X^{( l)})),
\label{eq:y}
\end{align}
\noindent
where $\text{LN}(\cdot)$ denotes the Layer Normalization operation.
The spatial path undergoes spatial attention, while the temporal path undergoes space-time attention following TimeSformer~\cite{bertasius2021space}. 

\subsubsection{Information exchange} 
As shown in \figref{overview} (a), each expert path exchanges information with other experts through the B-CA module.
We apply the B-CA operation $\Phi_{E_2 \rightarrow E_1}(\cdot)$ from the expert $E_2$ (key and value) to the expert $E_1$ (query) along with a skip connection to obtain information-fused features as follows: 
\begin{align} 
\B^{(l)}_{E_1} &= \Y^{(l)}_{E_1} + \Phi_{E_2 \rightarrow E_1}(\Y_{E_1}^{(l)}, \Y_{E_2}^{(l)}). 
\label{eq:b} 
\end{align} 
Similarly, we can apply the B-CA operation $\Phi_{E_1 \rightarrow E_2}(\Y_{E_2}^{(l)},\Y_{E_1}^{(l)})$ from the expert $E_1$ to the expert $E_2$ as well. 
Therefore, a B-CA module has two B-CA operations for bidirectional information exchange.
In this work, we configure the B-CA module with various expert pairs: audio \& spatial (A\&S), spatial \& temporal (S\&T), and audio \& temporal (A\&T).
For example, in the S\&T configuration, \ie CAST, we can employ a spatial expert as $E_1=S$ and a temporal expert as $E_2=T$, resulting in two B-CA operations:  $\Phi_{T\rightarrow S}(\cdot)$ and $\Phi_{S\rightarrow T}(\cdot)$.
We describe the B-CA operation $\Phi(\cdot)$ in detail in \secref{bcast}.

Finally, we feed the output $\B^{(l)}$ into a two-layer feed forward network (FFN)~\cite{dosovitskiy2020vit} with the GELU activation function in between the layers and another adapter to obtain the next layer input $\X^{(l+1)}$ along with a skip connection as follows:
\begin{align}
\X^{(l+1)}  &= \B^{(l)}  + \text{FFN}(\text{LN}(\B ^{(l)} )) + \text{ADAP}(\text{LN}(\B^{(l)} )).
\label{eq:next_x}
\end{align}
\noindent
Here, we drop the subscript $E_1$ or $E_2$ for brevity.

\subsubsection{Classification head}
To obtain the final prediction, we need to aggregate the outputs of both spatial and temporal experts. 
Specifically, for the spatial expert, we average the frame-level class tokens from the last attention block, $\X_{s}^{(12)}$, to obtain a single class token. We denote this operation as $\text{CLS}(\cdot)$.
To obtain temporal expert features, we aggregate all the tokens from the last attention block of the temporal expert, $\X_{t}^{(12)}$, using the global average pooling $\text{GAP}(\cdot)$ operation.
Then we add the adapter output of the CLS token and the adapter output of the GAP token to produce a fused token $\Z$:
\begin{align}
\label{eq:fuse}
\Z=\text{ADAP}(\text{CLS}(\X_S^{(12)}))+\text{ADAP}(\text{GAP}(\X_T^{(12)})).
\end{align}
We feed the fused token $\Z$ into a classification layer followed by the softmax function to obtain the predicted class probabilities. We train the model using the standard cross-entropy loss.

\vspace{-1em}
\subsection{Details of B-CA module}
\label{sec:bcast}
For brevity, we focus on describing the details of the S\&T B-CA operation among three types: \ie A\&S, S\&T, and A\&T. 

\subsubsection{Multi-Head Cross-Attention}
Multi-Head Cross-Attention (MHCA) is a variant of the MHSA operation~\eqref{eq:mhsa}, where query tokens come from one expert ($E_1$) and key and value tokens come from another expert ($E_2$). This allows the experts to exchange information and benefit from the strengths of each other. We define the MHCA operation as follows:
\begin{equation}
\label{eq:mhca}
\text{MHCA}(\Y_{E_1}, \Y_{E_2})=\text{Softmax}((\Y_{E_1} \mathbf{W}_{Q}) (\Y_{E_2} \mathbf{W}_{K})^{\top}) (\Y_{E_2} \mathbf{W}_{V}),
\end{equation}
where $\W_{Q}$, $\W_{K}$, and $\W_{V}$ are learnable query, key, and value parameter matrices respectively.
The S\&T B-CA module consists of bidirectional cross-attentions: Temporal-to-Spatial (T2S) and Spatial-to-Temporal (S2T). 

\subsubsection{Temporal-to-Spatial Cross-Attention}
In Temporal-to-Spatial (T$2$S) cross-attention, query tokens come from the spatial expert $S$, and key and value tokens come from the temporal expert $T$: $\text{MHCA}(\Y_S^{(l)}, \Y_T^{(l)})$.  We depict the attention window in \figref{overview} (c). Given a query, the model attends along the temporal dimension only. By using T2S cross-attention, the spatial expert can learn to attend to temporal features from the temporal expert. T2S MHCA leads to capturing spatio-temporal dependencies and improves the model performance.

\subsubsection{Spatial-to-Temporal Cross-Attention}
In Spatial-to-Temporal (S$2$T) cross-attention, query tokens come from the temporal expert $T$, and key and value tokens come from the spatial expert $S$: $\text{MHCA}(\Y_T^{(l)}, \Y_S^{(l)})$. We illustrate the attention window in \figref{overview} (c). Given a query, the model attends along the spatial dimension only. By using S2T cross-attention, the temporal expert can attend to fine-grained spatial features from the spatial expert. S2T MHCA leads to a more balanced spatio-temporal understanding and improves the performance in fine-grained action recognition.

\subsubsection{Bottleneck Cross-Attention in Space and Time}
To achieve efficient and effective learning, we incorporate the T2S and S2T MHCA into bottleneck-shaped adapters. We illustrate B-CA architecture in \figref{overview} (b). 
We plug the MHCA modules into adapters and add new learnable positional embeddings for each MHCA. We define the B-CA operation for T2S cross-attention $\Phi_{T\rightarrow S}(\cdot)$ as follows:
\begin{equation}
\label{eq:cast_bca}
\begin{aligned}
{\Y'_S}^{(l)} &= \E_S+\text{LN}(\Y_S^{(l)} \W_{D,S}),\\
\Phi_{T\rightarrow S}(\Y_S^{(l)}, \Y_T^{(l)}) &= \sigma(\text{MHCA}({\Y'_S}^{(l)},{\Y'_T}^{(l)})) \W_{U,S},
\end{aligned}
\end{equation}
where $\W_{D,S}$ and $\W_{U,S}$ are linear down- and up-projection matrices for the spatial expert $S$, $\E_S$ is a positional embedding for the spatial expert, and $\sigma(\cdot)$ is the GELU activation function, respectively.
We can define the B-CA operation for S2T cross-attention $\Phi_{S\rightarrow T}(\cdot)$ similarly. The output of B-CA goes into a feed forward network using \eqref{eq:next_x}.

\vspace{-1em}
\subsection{CAVA architecture} 
\label{sec:architecture_cava}
CAVA extends the CAST architecture by incorporating an audio expert, enabling effective audio-visual understanding. 
While the overall architecture is similar to CAST (\secref{architecture_cast}), CAVA is specifically designed to handle multi-modal inputs. 

\subsubsection{Audio-to-Spatial Cross-Attention}
In Audio-to-Spatial (A$2$S) cross-attention, query tokens come from the spatial expert $S$, while key and value tokens come from the audio expert $A$: $\text{MHCA}(\Y_S^{(l)}, \Y_A^{(l)})$. The A2S cross-attention uses a space-time attention window, called global attention, that attends to all tokens across both the spatial and temporal dimensions, following the global attention strategy of TimeSformer~\cite{bertasius2021space}. This approach is particularly suited for spectrograms, where both frequency and time information are contained within a single frame. By using global attention, we achieve better alignment between visual and audio patches suitable for spectrograms, leading to improved performance in audio classification tasks. 
The Spatial-to-Audio (S$2$A) cross-attention is similar, except that the query comes from the audio expert $A$ and the key and value come from the spatial expert $S$. We omit further details as the operations are symmetric.

\subsubsection{Time embedding}
\label{time_embedding}
To better align the spectrogram and visual modalities, we utilize the Time Interval MLP~\cite{chalk2024tim}. 
The Time Interval MLP takes the start and end time stamps of an interval as an input and outputs an encoding that captures both the relative position and duration of the interval. 
Unlike traditional position embeddings, the Time Interval MLP results in the embeddings of continuous time intervals, making it suitable for the audio modality.

In the spatial path, each frame out of total $t$ frames receives a unique time embedding $\text{\textbf{E}}^{time}_S$ that represents a non-overlapping time interval.
All patches within the same frame share the same embedding.
In the audio path, the input is a single spectrogram with both frequency and time dimensions. We divide the time axis into $t$ groups of patches, corresponding to the $t$ frames in the spatial path.
We define the B-CA operation for A2S cross-attention $\Phi_{A\rightarrow S}(\cdot)$ as follows:
\begin{equation}
\label{eq:cava_bca}
\begin{aligned}
{\Y'_S}^{(l)} &= \text{\textbf{E}}^{time}_S+\text{LN}(\Y_S^{(l)} \W_{D,S}),\\
\Phi_{A\rightarrow S}(\Y_S^{(l)}, \Y_A^{(l)}) &= \sigma(\text{MHCA}({\Y'_S}^{(l)},{\Y'_A}^{(l)})) \W_{U,S},
\end{aligned}
\end{equation}
\noindent We can define the B-CA operation for S2A cross-attention $\Phi_{S\rightarrow A}(\cdot)$ similarly.
By applying the Time Interval MLP~\cite{chalk2024tim}, we achieve improved alignment between the audio spectrogram features and visual features, as demonstrated in \tabref{cava_ablation:e}.

\vspace{-1em}
\subsection{\texorpdfstring{$\text{CA}^2\text{ST}$}{} architecture}
\label{sec:caast}
We extend CAVA by incorporating three expert models-audio, spatial, and temporal (A\&S\&T)-to achieve a more holistic audio-visual understanding of videos. 
\caast{} uses all three types of B-CA modules, A\&S, S\&T, A\&T, to facilitate information exchange between the multiple experts as illustrated in \figref{overview} (a).
\subsubsection{Information exchange between three experts}

Each path of \caast{} has two types of B-CA modules. We define the B-CA operation among them as follows:
\begin{align} 
\B^{(l)}_{E_1} &= \Y^{(l)}_{E_1} + \Phi_{E_2 \rightarrow E_1}(\Y_{E_1}^{(l)}, \Y_{E_2}^{(l)})+\Phi_{E_3 \rightarrow E_1}(\Y_{E_1}^{(l)}, \Y_{E_3}^{(l)}),
\label{eq:b_caast} 
\end{align} 
\noindent where $\Phi_{E_2 \rightarrow E_1}(\cdot)$ denotes the B-CA operation from the expert $E_1$ to the expert $E_2$. Each B-CA operation $\Phi(\cdot)$ uses the corresponding positional embeddings and the attention window shape specific to the B-CA type.     

\subsubsection{Parameter sharing in projection layers}
To improve the parameter efficiency of \caast{}, each expert uses the same parameters of the linear down-projection ($\W_D$) and up-projection ($\W_U$) within the B-CA module.
For example, given an expert $E_1$, both $\Phi_{E_2 \rightarrow E_1}$ and $\Phi_{E_3 \rightarrow E_1}$ operations share the same parameters $\W_U$ and $\W_D$ in \eqnref{b_caast}.
Different experts have distinct down- and up-projection parameters.
Due to the parameter sharing, \caast{} uses only 62M learnable parameters, despite employing three experts as shown in \tabref{main_audio:epicsounds}.

\vspace{\secmargin}
\section{Experimental Results}
\label{sec:results}

In this section, we present the experimental results that answer the following research questions regarding \red{our proposed models---CAST, CAVA, and \caast{}}: (1) 
Do existing methods show a balanced spatio-temporal understanding of videos? (\secref{balance}) 
(2) What are the ingredients for a balanced spatio-temporal understanding? (\secref{balance}) 
(3) Is the proposed method effective? (\secref{balance}, \secref{analysis}) 
(4) How can we effectively combine spatial and temporal models to achieve such balance? (\secref{ablation}) 
(5) Does the proposed method work across different modalities such as audio and visual? (\secref{cava_analysis}, \secref{cava_ablation}) 
(6) How can we effectively combine audio and video models? (\secref{cava_ablation}) 
(7) Does the proposed method outperform state-of-the-art methods in terms of balanced spatio-temporal understanding? (\secref{comparison}) 
(8) Does the proposed method outperform state-of-the-art methods in terms of holistic audio-video understanding? (\secref{cava_comparison},\secref{robustness})   
(9) Does the proposed method offer a favorable trade-off between performance and complexity? (\secref{complexity})
(10) How does B-CA module enable effective semantic transfer across modalities? (\secref{analysis_bca})
To this end, we first provide details about the datasets and implementation in \secref{dataset} and \secref{imple}, respectively.

\vspace{-0.4em}

\vspace{\secmargin}
\subsection{Datasets} 
\label{sec:dataset}

\subsubsection{Action recognition.} 
We evaluate the CAST on two public datasets for conventional action recognition: Something-Something-V2 (SSV2)~\cite{goyal2017something} and Kinetics-400 (K400)~\cite{kay2017kinetics}. 
The SSV2 requires more temporal reasoning~\cite{bertasius2021space,kowal2022deeper} while the K400 is relatively static biased~\cite{li2018resound,Choi-NeurIPS-2019,sevilla2021only}.

\subsubsection{Fine-grained action recognition.}
We evaluate the CAST on the fine-grained action recognition task: EPIC-KITCHENS-100 (EK100)~\cite{damen2022rescaling}.
In contrast to conventional action recognition, EK100 defines an action as a combination of a verb and a noun. Therefore, we refer to the action recognition in EK100 as \emph{fine-grained action recognition}. Since fine-grained action recognition requires correctly predicting both the verb and the noun to recognize an action it is more challenging than conventional action recognition, which requires predicting a single action label: \eg K400 or SSV2.
Additionally, EK100 provides participant IDs, enabling an evaluation setting where the model is tested on participants not seen during training—referred to as the \emph{cross-subject setting}.
The unseen participant setting is more challenging than the conventional EK100 setting due to the domain shift across participants.
\subsubsection{Audio-visual recognition.}
We evaluate the CAVA and \caast{} on four public audio-visual recognition datasets: EPIC-SOUNDS~\cite{huh2023epic}, UCF-101~\cite{soomro2012dataset}, VGG-Sound~\cite{chen2020vggsound} and KineticsSound~\cite{xiao2020audiovisual}. In contrast to action recognition datasets, these datasets require models to effectively utilize audio information to accurately recognize actions. Particularly, EPIC-SOUNDS is challenging due to its egocentric perspective and the need to understand fine-grained interactions within a kitchen environment, unlike other audio-visual recognition datasets.

\subsubsection{Cross-domain benchmarks.}
To evaluate the generalization ability of our model, we use HD-EPIC~\cite{hd-epic} as a cross-domain validation benchmark. 
HD-EPIC consists of egocentric videos captured in-the-wild across diverse home environments.
We use HD-EPIC solely for validation, as no training data is provided.
Since HD-EPIC shares a subset of action classes with EK100, it enables cross-domain evaluation under the same label space. 
For the audio modality, we use the sound subset of HD-EPIC, which we refer to as \emph{HD-EPIC-SOUNDS}.
\subsubsection{Long‑video robustness.}
We evaluate CAST on the ActivityNet~\cite{caba2015activitynet} dataset, whose 65-second long videos of everyday activities provide a longer and noisy temporal context.

\begin{figure*}[!t]

    \centering
    \includegraphics[width=0.9\linewidth]{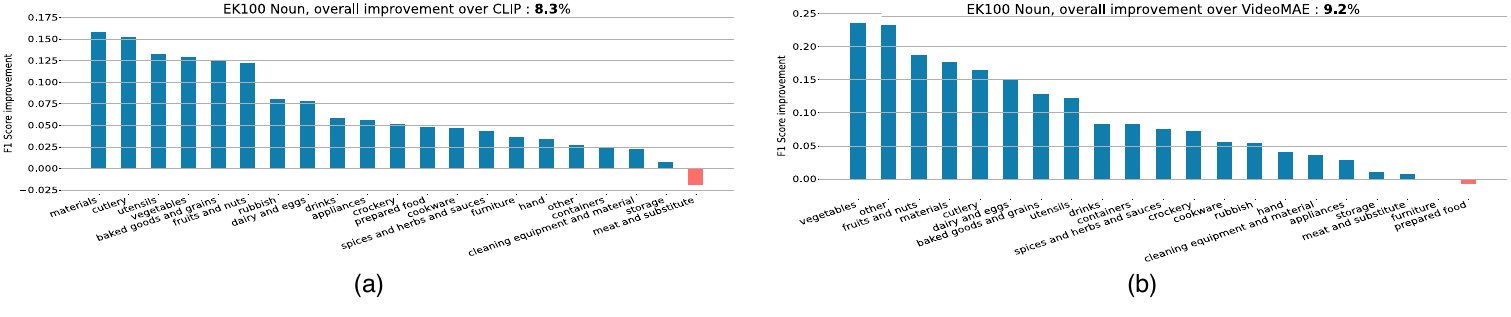}

    \vspace{-1em}
    \caption{\textbf{Improvements of CAST over each expert on EK100 noun classes.}  
    (a) Improvement over CLIP. CAST outperforms CLIP for every super-category except \emph{meat and substitute}. (b) Improvement over VideoMAE. CAST outperforms VideoMAE for every super-category except \emph{furniture} and \emph{prepared food}. 
    }
    \vspace{\figcapmargin}
    \label{fig:category-noun}
\end{figure*}
    
\begin{figure}[!t]    
    \centering    \includegraphics[width=0.8\columnwidth]{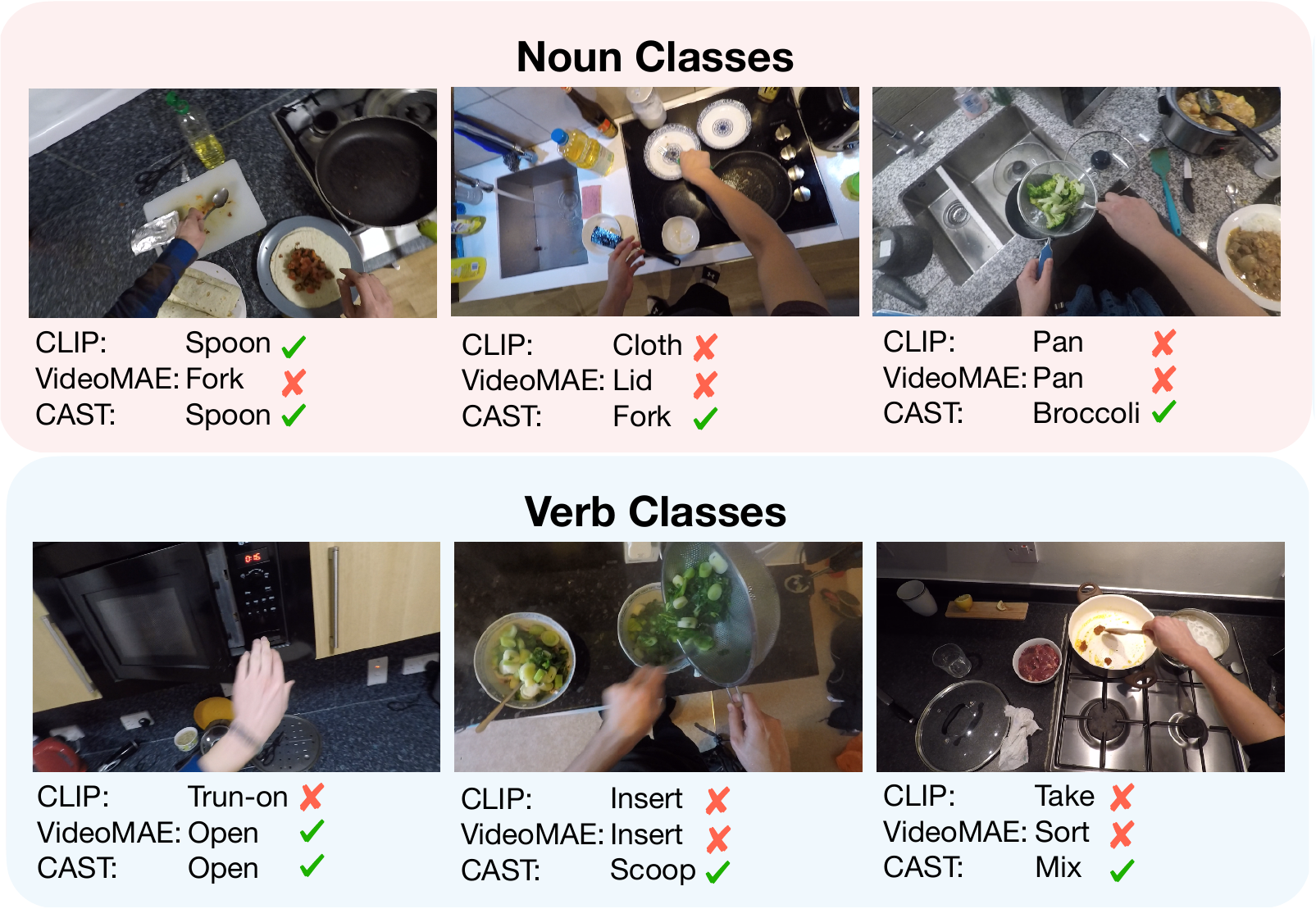}
    \figcaption{Qualitative examples from EK100 comparing CLIP, VideoMAE, and the proposed CAST}{Each expert model shows more accurate predictions in their expertise, but shows weaker performance on the other task. However, CAST consistently shows correct predictions for both tasks, demonstrating the effectiveness of the proposed B-CA module.
    }
    \vspace{-1.5em}
    \label{fig:qualitative}
\end{figure}

\vspace{\secmargin}
\subsection{Implementation details} 
\label{sec:imple}

In this section, we briefly provide our experimental setup and implementation details. 
Please refer to the supplementary materials for complete implementation details.
We conduct all the experiments with 16 NVIDIA GeForce RTX 3090 GPUs. We implement proposed methods using PyTorch and build upon the existing codebase of VideoMAE~\cite{tong2022videomae}. 
We sample 16 frames from each video to construct an input clip. For the K400 and all audio-visual recognition datasets, we apply dense sampling~\cite{feichtenhofer2019slowfast}, while for SSV2, EK100 and EPIC-SOUNDS, we use uniform sampling~\cite{wang2018temporal}.
We then perform random cropping and resizing every frame into $224\times224$ pixels.
We use the AdamW~\cite{loshchilov2017decoupled} optimizer with momentum betas of (0.9, 0.999)~\cite{chen2020generative} and a weight decay of 0.05. By default, we train the model for 50 epochs, with the cosine annealing learning rate scheduling~\cite{loshchilov2016sgdr} and a warm-up period of 5 epochs. The default base learning rate, layer decay~\cite{bao2021beit}, and drop path are set to 0.001, 0.8, and 0.2, respectively. We freeze all the parameters of each expert, except for the B-CA modules, adapters, and the last layer normalization. We set the batch size per GPU as 6 with update frequency of 2 for CAST and 4 for \caast{}.

\subsubsection{Inference} 
Given an input video, we randomly sample frames multiple times to construct input clips with multiple temporal views with multiple spatial crops. After the temporal frame sampling, we resize every frame so that the shorter side has $224$ pixels. Then we perform spatial cropping to get multiple $224\times224$ crops for each clip. The final prediction is the average over (temporal views) $\times$ (spatial crops).
We use (5 clips) $\times$ (3 crops) views for the K400 and KineticsSound and (2 clips) $\times$ (3 crops) for the other datasets.

\vspace{\secmargin}
\subsection{Balanced spatio-temporal understanding} 
\label{sec:balance}

In \tabref{main_rgb}, we present the top-1 accuracies of several existing models. 
In the EK100 verb prediction task, VideoMAE outperforms ST-Adapter with a margin of $2.9$ points ($70.5\%$ \vs $67.6\%$), while in the EK100 noun prediction task, ST-Adapter~\cite{pan2022st} outperforms VideoMAE~\cite{tong2022videomae} with a margin of $3.6$ points ($55.0\%$ \vs $51.4\%$). 
Similarly, BEVT~\cite{wang2022bevt} outperforms AIM~\cite{yang2023aim} with a margin of $2.5$ points ($70.6\%$ \vs $68.1\%$) on the SSV2 dataset, whereas AIM outperforms BEVT with a margin of $3.9$ points ($84.5\%$ \vs $80.6\%$) on the K400 dataset. We observe similar trends for other methods as well. 
Our findings indicate that many existing models tend to exhibit a significant imbalance in their spatial or temporal understanding.

\subsubsection{Ingredients for balanced spatio-temporal understanding} 
To achieve a more balanced spatio-temporal understanding, we can employ two expert models: a spatial expert and a temporal expert. For the spatial expert, we use CLIP~\cite{radford2021learning}, which has demonstrated impressive performance on various image-based computer vision tasks. For the temporal expert, we use VideoMAE~\cite{tong2022videomae}, which has shown favorable performance on temporal-biased tasks such as SSV2 and EK100 verb prediction tasks. (Please refer to \secref{comparison} for the accuracy details.) While each expert is highly specialized in its own domain, we aim to create synergy between them by exchanging information to achieve the balanced spatio-temporal understanding.

\subsubsection{Effect of CAST} 

In \tabref{main_rgb}, we report the harmonic mean of top-1 accuracies for EK100 noun, EK100 verb, SSV2, and K400 across several models. The harmonic mean is an effective metric for assessing balanced spatio-temporal understanding performance, as it gives more weight to lower-performing tasks. A higher harmonic mean value indicates that the performance over the different tasks is more balanced. Our spatial expert, CLIP achieves an accuracy of $56.5\%$, while the temporal expert, VideoMAE achieves $66.6\%$, and our CAST achieves $71.6\%$. These results validate the effectiveness of our CAST, which enables the spatial and temporal experts to collaborate and make synergistic predictions by exchanging information with each other through cross-attention.

\begin{table}[t]
\centering
\captionsetup{justification=justified}

\newcommand{\cmark}{\ding{51}}%
\newcommand{\xmark}{\ding{55}}%
\caption{\tb{Effect of information exchange.}}
\resizebox{0.75\linewidth}{!}{
\begin{tabular}{lcccc}
\toprule    
&& \multicolumn{3}{c}{Top-1 Acc.}\\
\cmidrule(lr) {3-5}
Method && Verb & Noun & Act.\\
\midrule
Indep. experts w/o adapter && 70.7 & 50.1 &  40.0 \\
Indep. experts w/ adapter && 68.1 & 54.2 & 41.7 \\
Ensemble of experts w/ adapter && 68.2 & 55.3 & 42.9 \\
CAST && \cellcolor{gray!30}\textbf{72.5} & \cellcolor{gray!30}\textbf{60.3} & \cellcolor{gray!30}\textbf{48.7}\\
\bottomrule
\end{tabular}
}
\label{tab:ablation:a}
\vspace{-0.5em}
\end{table}
\vspace{\secmargin}
\subsection{Analysis on fine-grained action recognition}
\label{sec:analysis}

In this section, we provide a detailed analysis of how CAST improves the balanced spatio-temporal understanding in the fine-grained action recognition task: EK100.

\subsubsection{Category-level performance analysis}
In \figref{category-noun}, We present the EK100 noun super-category-wise weighted average F1 score improvement of CAST over our spatial expert (CLIP) and temporal expert (VideoMAE). In \figref{category-noun} (a), we observe that CAST significantly improves upon the spatial expert, CLIP, in several super-categories such as \emph{cutlery}, \emph{utensils}, and \emph{vegetables}. 
Similarly, in \figref{category-noun} (b), we observe that CAST significantly improves upon the temporal expert, VideoMAE, in several categories such as \emph{vegetables} and \emph{cutlery}.
These results indicate that CAST achieves a more accurate understanding of fine-grained small objects by leveraging temporal context from the temporal expert and spatial context from CLIP.

\subsubsection{Qualitative analysis}
To better understand the effectiveness of CAST, we provide qualitative analysis on a few sample frames from the EK100 dataset in \figref{qualitative}. We show the predictions of CLIP, VideoMAE and CAST. While each expert model performs well their respective tasks of expertise, it struggles in others. In contrast, CAST consistently makes correct predictions for both noun and verb tasks, such as \emph{spoon} and \emph{open}, demonstrating its balanced spatio-temporal understanding for fine-grained action recognition.

\vspace{\secmargin}
\subsection{Ablation study on CAST architecture}
\label{sec:ablation}
\begin{table}[t]
\centering
\captionsetup{justification=justified}

\newcommand{\cmark}{\ding{51}}%
\newcommand{\xmark}{\ding{55}}%
\caption{\tb{Different information exchange methods.}}
\resizebox{0.75\linewidth}{!}{
\begin{tabular}{lccccc}
\toprule
 & & & \multicolumn{3}{c}{Top-1 Acc.}\\
\cmidrule(lr) {4-6}
Method & Late & Layer-wise & Verb & Noun & Act.\\
\midrule
Add & \cmark & & 68.9 & 56.6 & 44.2 \\
Concat & \cmark & & 69.2 & 56.4 & 44.5 \\
Lateral & &\cmark & 68.9 & 49.1 & 39.0 \\
CAST & &\cmark & \cellcolor{gray!30}\textbf{72.5} & \cellcolor{gray!30}\textbf{60.3}& \cellcolor{gray!30}\textbf{48.7}\\
\bottomrule
\end{tabular}
}
\label{tab:ablation:b}

\end{table}

\begin{table}[t]
\centering
\captionsetup{justification=justified}

\newcommand{\cmark}{\ding{51}}%
\newcommand{\xmark}{\ding{55}}%
\caption{\tb{Design choice of B-CA module.}}
\resizebox{0.72\linewidth}{!}{
\begin{tabular}{lccccc}
\toprule
& Tune && \multicolumn{3}{c}{Top-1 Acc.}\\
\cmidrule(lr) {3-6}
Method & Param(M) && Verb & Noun & Act.\\
\midrule
Identity & 18.1 && 68.1 & 54.2 & 41.7\\
w/o adapter & 85.9 && 69.3 & 49.4 & 39.4\\
X-attn.$\rightarrow$adapter & 93.0 && 71.3 & 60.1 & 47.9\\
B-CA & 44.8 && \cellcolor{gray!30}\textbf{72.5} & \cellcolor{gray!30}\textbf{60.3} & \cellcolor{gray!30}\textbf{48.7}\\
\bottomrule
\end{tabular}
}
\label{tab:ablation:c}
\vspace{-0.5em}
\end{table}
\begin{table}[t]
\centering
\captionsetup{justification=justified}

\newcommand{\cmark}{\ding{51}}%
\newcommand{\xmark}{\ding{55}}%
\caption{\tb{Effect of cross-attention window shape.}}
\resizebox{0.72\linewidth}{!}{
\begin{tabular}{cccccc}
\toprule
\multicolumn{2}{c}{Window shape} && \multicolumn{3}{c}{Top-1 Acc.}\\
\cmidrule(lr){1-2} \cmidrule{4-6}
T2S & S2T && Verb & Noun & Act.\\
\midrule
space-time & space-time && 71.0 & 59.3 & 47.2\\
space-time & space && 71.9 & 60.3 & 48.4 \\
space & space && 72.3 & 60.2 & 48.5 \\
time & space && \cellcolor{gray!30}\textbf{72.5} & \cellcolor{gray!30}\textbf{60.3} & \cellcolor{gray!30}\textbf{48.7} \\
\bottomrule
\end{tabular}
}
\label{tab:ablation:e}
\end{table}
\begin{table}[t]
\centering

\newcommand{\cmark}{\ding{51}}%
\newcommand{\xmark}{\ding{55}}%
    \caption{\tb{Effect of bi-directional cross-attention.}}
    \resizebox{0.7\linewidth}{!}{
        \begin{tabular}{lcccc}
        \toprule    
        & & \multicolumn{3}{c}{Top-1 Acc.}\\
        \cmidrule(lr) {3-5}
        Method && Verb & Noun & Act.\\
        \midrule
        Indep. experts w/ adapter && 68.1 & 54.2 & 41.7 \\
        S2T only && 71.2 & 55.0 &  43.7 \\
        T2S only && 68.7 & 60.5 & 46.7 \\
        CAST && \cellcolor{gray!30}\textbf{72.5} & \cellcolor{gray!30}\textbf{60.3} & \cellcolor{gray!30}\textbf{48.7}\\
        \bottomrule

\end{tabular}
}
\label{tab:ablation:g}
\vspace{-0.5em}

\end{table}
\begin{table}[t]
\centering
\captionsetup{justification=justified}

\newcommand{\cmark}{\ding{51}}%
\newcommand{\xmark}{\ding{55}}%
    \caption{\tb{Role of each expert.}}
    \resizebox{0.75\linewidth}{!}{
        \begin{tabular}{cccccc}
        \toprule
        \multicolumn{2}{c}{Expert} && \multicolumn{3}{c}{Top-1 Acc.}\\
        \cmidrule(lr){1-2} \cmidrule{4-6}
        Spatial & Temporal  && Verb & Noun & Act.\\
        \midrule
        CLIP~\cite{radford2021learning} & CLIP && 69.3 & 58.8 & 46.0\\
        VideoMAE~\cite{tong2022videomae} & CLIP && 72.2 & 58.8 & 47.8\\
        VideoMAE & VideoMAE && 69.8 & 49.9 & 40.3\\
        CLIP & VideoMAE && \cellcolor{gray!30}\textbf{72.5} & \cellcolor{gray!30}\textbf{60.3} & \cellcolor{gray!30}\textbf{48.7} \\
        \bottomrule

\end{tabular}
}
\label{tab:ablation:h}
\end{table}

We conduct comprehensive ablation studies to examine the design choices for the proposed CAST and B-CA architecture. 
Here we conduct all experiments on the EK100~\cite{damen2022rescaling} dataset with 16-frame input videos and report the top-1 accuracy on the validation set. 
We employ CLIP~\cite{radford2021learning} as a spatial expert and VideoMAE-B/16~\cite{tong2022videomae} as a temporal expert.
\red{Unless otherwise specified, we refer to VideoMAE-B/16 simply as VideoMAE.}
For a fair ablation study, we use the same hyperparameters for each experiment unless explicitly mentioned. 

\subsubsection{Effect of information exchange}

We investigate whether CAST effectively achieves a synergistic effect by exchanging information between the two expert models. In \tabref{ablation:a}, we compare CAST with three baselines. i) A baseline using two independent expert models without any information exchange (fully fine-tuned). ii) The same baseline as i), but we add adapters and fine-tune the adapters and the classification heads only, iii) A test-time ensemble of two independent experts (with adapters and heads fine-tuning only).
In these baselines, the spatial model predicts nouns and the temporal model predicts verbs. We observe that ensembling the two expert models improves action accuracy by at least $1.2$ points compared to the baselines without information exchange. Furthermore, CAST achieves the highest action accuracy of 48.7\%, indicating that information exchange is essential for achieving balanced spatio-temporal understanding.

\subsubsection{Comparison with simple information exchange baselines}

We compare CAST with simple information exchange baselines: i) late fusion with addition, ii) late fusion with concatenation, iii) layer-wise fusion using the bidirectional lateral connection (element-wise addition) with a linear projection layer.  
We fine-tune the adapters and the classification heads only in all three baselines.
In the case of late fusion, we put the information exchange in the penultimate layer.
We present the results in \tabref{ablation:b}. Notably, the late fusion baselines cause a significant drop in performance. Additionally, we observe that layer-wise fusion without cross-attention underperforms the simple late fusion baselines. These results highlight the importance of cross-attention for enabling effective information exchange between spatial and temporal experts.

\subsubsection{Design of B-CA module}

To investigate the most effective design for the B-CA module in facilitating information exchange between the two expert models, we conduct an ablation study and present the results in \tabref{ablation:c}.
The first row shows a baseline without the B-CA module, which is equivalent to the identity function. Compared to this baseline, B-CA achieves a significant improvement of 7.0 points in action accuracy, highlighting the importance of information exchange between two experts.    
The second row represents a baseline where the bottleneck adapters (\eg down- and up-projection) are removed from the B-CA module. The 9.3-point gap between this baseline and B-CA underscores the importance of bottleneck adapters for effective information exchange between the two expert models.
The third row (\textit{X-attn.$\rightarrow$adapter}) represents a baseline with the adapters after cross-attention. Compared to B-CA, this baseline shows a $0.8$ points drop in action accuracy while having more than double the number of learnable parameters ($44.8M$ \vs $93.0M$). 
The results show that placing cross-attention within the bottleneck is more effective and efficient than other baselines. In summary, placing cross-attention in the middle of the bottleneck architecture allows B-CA to enable effective information exchange between the two experts, leading to a synergistic effect and fewer learnable parameters.

\subsubsection{Effect of cross-attention window shape}
We investigate how the window shape in the cross-attention mechanism affects the T2S and S2T modules, as shown in \tabref{ablation:e}.
Please refer to \figref{overview} (c) for the details of the window shape. 
Using space-time attention for both T2S and S2T modules results in the worst performance. We conjecture that learning joint space-time attention is challenging with the given model capacity~\cite{bertasius2021space}. In contrast, using time attention in T2S and space attention in S2T achieves the best performance. As a result, we apply this configuration of window shape to our CAST by default.

\subsubsection{Effect of bi-directional cross-attention}

As depicted in \figref{overview} (b), B-CA facilitates bi-directional cross-attention for information exchange between experts.
To validate the effectiveness of this bi-directional information exchange, we compare CAST with unidirectional information exchange baselines equipped with S2T or T2S cross-attention only. Each unidirectional information exchange baseline still has both experts. 
In \tabref{ablation:g}, compared to our CAST (48.7\%), the S2T and T2S-only baselines show drops of 5.0 and 2.0 points in accuracy, respectively. The results validate the effectiveness of the proposed bi-directional cross-attention.

\subsubsection{Role of each expert}
In \tabref{ablation:h}, we investigate the role of experts within CAST by controlling the role assignment to each expert. 
We observe that we can achieve the best action accuracy of $48.7\%$ when we employ CLIP as our spatial expert and VideoMAE as our temporal expert, as originally used in CAST. When we use VideoMAE as both the spatial and temporal expert, we achieve an accuracy of $40.3\%$, while using CLIP in both roles results in $46.0\%$ in action accuracy.
Interestingly, as shown in the second row of the table, when we revert the role of CLIP and VideoMAE, we achieve a good performance of $47.8\%$. The results demonstrate that the B-CA architecture facilitates effective information exchange between the two experts. Through the stacked B-CA, the experts can learn high-quality spatio-temporal representations by exchanging information, even when the roles are reverted.
In summary, these findings suggest that CAST achieves optimal performance when we assign models to roles that align with their strengths. CLIP serves as an effective spatial expert, whereas VideoMAE is more effective as a temporal expert. The B-CA architecture enables these experts to leverage their respective strengths through information exchange, leading to enhanced balanced spatio-temporal understanding.
\begin{table}[t]
    \centering
    \caption{\tb{CAST with CNN-based expert.}
    }
    \resizebox{0.8\linewidth}{!}{
        \begin{tabular}{lc ccc}
        \toprule
         & & \multicolumn{3}{c}{Top-1 Acc. } \\
         \cmidrule(lr) {3-5}
         Method & Backbone & Verb & Noun & Act. \\
         \midrule
         ResNet-50~\cite{He2015DeepRL} & - & 43.6 & 43.6 & 24.4 \\
         TSN~\cite{wang2018temporal} & ResNet-50&60.2& 46.0& 33.2 \\ 
         TRN~\cite{zhou2018temporal} & ResNet-50&65.9&45.4&35.3 \\
         TSM~\cite{lin2019tsm} & ResNet-50&67.8&49.0&38.3 \\
         SlowFast~\cite{feichtenhofer2019slowfast} & ResNet-50 & 54.9 & 50.0 & 38.5 \\
         \midrule
         VideoMAE~\cite{tong2022videomae} & ViT-B & 70.5 & 51.4 & 41.7 \\
         \midrule
         CAST & ResNet-50\&ViT-B &  \cellcolor{gray!30}{\tb{71.4}} &  \cellcolor{gray!30}{\tb{53.9}} &  \cellcolor{gray!30}{\tb{43.8}}\\
        \bottomrule
        \end{tabular}
    }
    \label{tab:cnn_extend}
    \vspace{-1em}
\end{table}

\subsubsection{Effect of expert model architecture}
To explore the applicability of the B-CA module beyond transformer-based architectures, we employ ResNet-50~\cite{He2015DeepRL}, pretrained on ImageNet-1K, as the spatial expert and VideoMAE~\cite{tong2022videomae} as the temporal expert.
We insert a BC-A module after each ResNet-50 stage to exchange information with VideoMAE.
\tabref{cnn_extend} shows CAST performance with other baseline models using ResNet-50 backbone.
CAST outperforms both expert models—ResNet-50 (24.4\%) and VideoMAE-B (41.7\%)—achieving 43.8\% action accuracy, validating the effectiveness of our B-CA module.
For further details of the hybrid setup, please refer to Section~\alwaysred{I-E} of the supplementary material.
\begin{figure*}[t]

    \centering
    \includegraphics[width=0.9\linewidth]{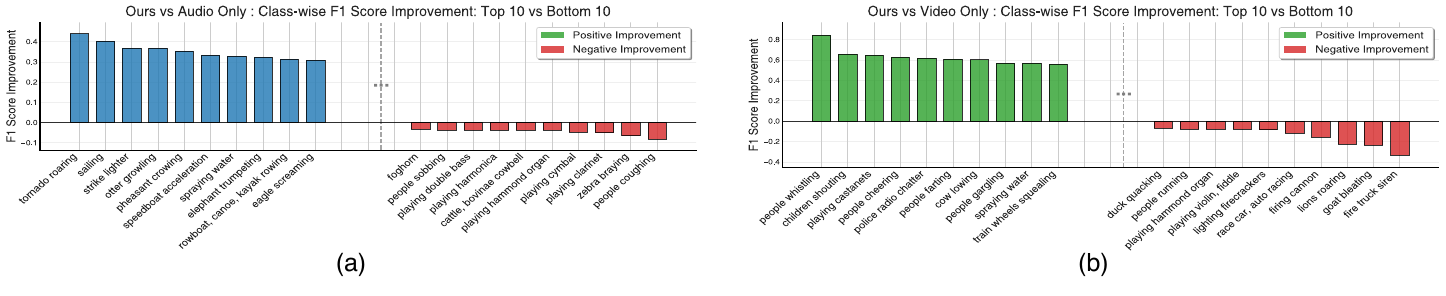}
    
    \vspace{-1em}
    \caption{\textbf{Class-wise F1 score improvements of CAVA over its expert models on the audio-visual classes of the EPIC-SOUNDS dataset.} 
    Each figure highlights the top-10 classes with the most significant gains and the bottom-10 with the least, clarifying our model's strengths and limitations.
    (a) Comparison with the audio expert (AST).
    (b) Comparison with the visual expert (CLIP).} 
    \label{fig:ast_clip_cava_f1}
    \vspace{-1em}
\end{figure*}

\begin{figure}[!t]
\centering
\includegraphics[width=0.7\columnwidth]{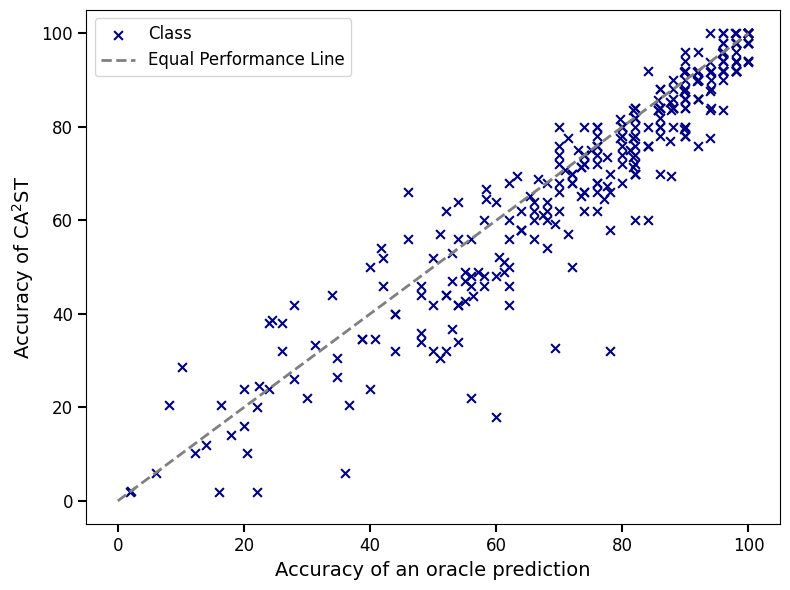}
\figcaption{Class-wise performance comparison between \caast{} and an oracle on the VGG-Sound~\cite{chen2020vggsound} dataset}
{The horizontal axis shows the class-wise accuracy of an oracle of CAST as the visual expert model and AST as the audio expert model, while the vertical axis shows the class-wise accuracy of \caast{}. The dashed line indicates equal performance. The points above the dashed line are classes where \caast{} outperforms the oracle.}

\vspace{-0.5em}
\label{fig:vgg_class-wise}
\end{figure}

\vspace{\secmargin}
\subsection{Synergy from audio and visual models}
\label{sec:cava_analysis}

Here, we explore whether CAVA and \caast{}, our audio-visual models, create synergy between audio and visual models.
In Figure~\ref{fig:ast_clip_cava_f1}, we show the class-wise F1 score improvements achieved by CAVA over the modality-specific experts (AST and CLIP), highlighting how the B-CA module enables effective integration of complementary information.
In (a), CAVA shows significant gains over the audio expert (AST) for classes where visual context helps disambiguate similar sounds, \eg a funnel cloud in \emph{tornado roaring} could be helpful.
However, for visually ambiguous classes such as \emph{playing clarinet}, performance may slightly degrade as visual confusion can override the clear audio signal.
In (b), CAVA improves over the visual expert (CLIP) for audio-centric classes like \emph{people whistling} and \emph{people gargling}, where visual features may be ambiguous.
Conversely, in visually distinctive classes such as \emph{race car}, adding audio could slightly degrade the performance.
These results confirm that our B-CA module is most effective when modalities contribute complementary, non-redundant information.

In \figref{vgg_class-wise}, we present the class-wise accuracy of \caast{} and an oracle on the VGG-Sound~\cite{chen2020vggsound} dataset.
To calculate the oracle performance, we use two models: i) CAST as the visual model and ii) AST as the audio model.
For each video, both models predict an action, and we count the prediction as correct if either the visual or audio model correctly identifies the action.
The dashed line represents equal performance. 
Out of 309 classes, 188 classes fall within a margin of $5.0$ points of the line, with 107 classes above the line, indicating that \caast{} either matches or outperforms the oracle.
The result underscores the effectiveness of \caast{} in facilitating information exchange between modalities, leading to creating synergy between the audio and visual expert models.

\vspace{\secmargin}
\subsection{Ablation study on CAVA}
\label{sec:cava_ablation}
\begin{table}[t]
\centering
\captionsetup{justification=justified}
    \caption{\tb{Effect of information exchange.}}
        \resizebox{0.75\linewidth}{!}
        {
            \begin{tabular}{lcc}
            \toprule    
            Method && Top-1 Acc.\\
            \midrule
            Audio expert model only && 54.9  \\
            Visual expert model only && 31.9  \\
            Ensemble of audio-visual experts && 55.6\\
            CAVA && \cellcolor{gray!30}\textbf{60.3} \\
            \bottomrule
            \end{tabular}
        }
    \label{tab:cava_ablation:a}

    \vspace{-0.5em}

\end{table}
\begin{table}[t]
\centering
\captionsetup{justification=justified}
    \caption{\tb{Different information exchange methods.}}
        \resizebox{0.65\linewidth}{!}{
            \begin{tabular}{lcc}
            \toprule    
            Method && Top-1 Acc.\\
            \midrule
            Late fusion w/ adapter && 57.9 \\
            Layer-wise connection && 55.2 \\
            
            B-CA && \cellcolor{gray!30}\textbf{60.3} \\
            \bottomrule
            \end{tabular}
}
    \label{tab:cava_ablation:b}
    \vspace{-0.5em}
\end{table}
\begin{table}[t]
\centering
\captionsetup{justification=justified}
\newcommand{\cmark}{\ding{51}}%
\newcommand{\xmark}{\ding{55}}%
    \caption{\tb{Effect of expert choice in CAVA.}}
    \resizebox{0.55\linewidth}{!}{ 
        \begin{tabular}{lcccc}
        \toprule
         \multicolumn{3}{c}{Expert} & \multirow{2}{*}{Top-1 Acc.}  \\
         \cmidrule(lr){1-3} 
        Audio & Spatial & Temporal &  \\
        \midrule
         \cmark&  &  &\red{54.9}  \\
         & \cmark &  &\red{31.9}  \\
         &  & \cmark & \red{31.8} \\
         & \cmark & \cmark & 47.8  \\
        \cmark &  & \cmark & 58.8  \\
        \cmark & \cmark &  & \cellcolor{gray!30}60.3\\
        \cmark & \cmark & \cmark & \cellcolor{gray!30}\textbf{61.0} \\
        \bottomrule
        \end{tabular}
    }
    \label{tab:cava_ablation:c}
    \vspace{-0.5em}

\end{table}
\begin{table}[t]
\centering
\captionsetup{justification=justified}
    \caption{\tb{Effect of audio expert model.}}
    \resizebox{0.75\linewidth}{!}{
        \begin{tabular}{lcc}
        \toprule
         Pre-training data &Audio expert  & {Top-1 Acc.} \\
        \midrule
         \multirow{2}{*}{Image} &ViT-B/16 IN-1K~\cite{dosovitskiy2020vit}  & 55.6 \\
         &CLIP~\cite{radford2021learning}  & 56.8 \\
         \midrule
         \multirow{2}{*}{Audio spectrogram} &SSAST~\cite{gong2022ssast} & 59.2\\
         & AST~\cite{gong21b_interspeech}  & \cellcolor{gray!30}\textbf{60.3}\\
        \bottomrule
        \end{tabular}
        }
    \label{tab:cava_ablation:d}
    \vspace{-0.5em}
\end{table}
We conduct comprehensive ablation studies to validate our design choices for our proposed CAVA. 
Here we conduct all experiments on the EPIC-SOUNDS~\cite{huh2023epic} dataset, using 16-frame input videos and corresponding audio spectrograms and report top-1 accuracy on the validation set.
We employ AST~\cite{gong21b_interspeech} and CLIP~\cite{radford2021learning} as a audio and spatial expert model for CAVA.
Additionally, we employ VideoMAE~\cite{tong2022videomae} as a temporal expert model for \caast{}.
For a fair ablation study, we use same hyperparameters and expert models for each experiment unless explicitly mentioned.

\subsubsection{Effect of information exchange}
In \tabref{cava_ablation:a}, we investigate whether CAVA effectively achieves a synergistic audio-visual understanding by using both modalities.
We compare CAVA with three baselines: i) an audio expert model only and ii) a visual expert model only without any information exchange, iii) a test-time logit ensemble of two independent experts.
For a fair comparison, both audio and visual expert are frozen, and only the adapters are learned during training in all baselines.
We observe that the logit ensemble baseline shows an improvement in $0.7$ points compared to the audio expert model only baseline.
With information exchange, our proposed CAVA achieves the best accuracy of $60.3\%$ with a margin of $4.7$ points compared to the logit ensemble baseline. 
The results show that information exchange between two modalities is effective for achieving audio-visual understanding.

\subsubsection{Comparison with simple information exchange baselines}
We investigate whether the B-CA module outperforms other simple information exchange methods. 
In \tabref{cava_ablation:b}, we compare CAVA with two baselines: i) late fusion (addition) with adapters and ii) layer-wise fusion using bidirectional lateral connections (element-wise addition) with a linear projection layer. 
Despite having layer-wise connections, the linear projection method shows a $3.7$ points drop in accuracy compared to the late fusion method. 
In contrast, B-CA achieves the highest performance with an accuracy of 60.3\% with layer-wise cross-attention. 
The results demonstrate that CAVA, utilizing the B-CA module, is more effective in achieving audio-visual understanding compared to simple information exchange baselines.

\subsubsection{Expert choice for CAVA}
To investigate which expert combinations are most effective for audio-visual understanding, we compare the performance of different expert combinations for CAVA.
In \tabref{cava_ablation:c}, the spatial and temporal combination, which is equivalent to CAST, shows the lowest performance, highlighting the importance of the audio expert in achieving effective audio-visual understanding. 
As expected, we observe that the audio and spatial combination outperforms the audio and temporal combination.
Since audio spectrograms contain temporal information, an audio expert can create more synergy with a spatial expert rather than a temporal expert.
Therefore, we employ a spatial expert as a visual expert for CAVA in this paper by default. 
We achieve the best performance $61.0\%$ when we employ all three experts (audio, spatial, and temporal). 
We refer to this configuration as \caast{}, as depicted in \figref{overview}. 
The results shows that \caast{}, by adding a temporal expert, has a more holistic video recognition capability. 
\begin{table}[t]
\centering
\captionsetup{justification=justified}

    \caption{\tb{Effect of positional embedding methods.}}
    \resizebox{0.7\linewidth}{!}{
        \begin{tabular}{lcc}
        \toprule
      {Method} & {Param(M)} & {Top-1 Acc.} \\
        \midrule
        No Positional Embedding & 43.2 & 59.5  \\
        Positional Embedding~\cite{dosovitskiy2020vit} & 62.0 & 59.7   \\
        Time Embedding~\cite{chalk2024tim} & 43.8 & \cellcolor{gray!30}\textbf{60.3} \\ 
        \bottomrule 
        \end{tabular}
    }
    \label{tab:cava_ablation:e}
    \vspace{-0.5em}
\end{table}

\subsubsection{Effect of employing different models as audio experts} 
To explore the impact of using different audio expert models in CAVA, we compare the performance of employing different audio models as an audio expert while fixing the spatial expert as CLIP in CAVA. 
In \tabref{cava_ablation:d}, we evaluate two image pre-trained models, IN-1K pre-trained ViT-B/16 (IN-1K)~\cite{dosovitskiy2020vit} and CLIP~\cite{radford2021learning} as well as two audio spectrogram pre-trained models, AST~\cite{gong21b_interspeech} and SSAST~\cite{gong2022ssast}. 
We report the top-1 accuracy for model on the EPIC-SOUNDS~\cite{huh2023epic} dataset.
As expected, using audio pre-trained models outperform using image pre-trained models as the audio expert.
Employing AST as the audio expert achieves the highest accuracy of 60.3\%, outperforming image pre-trained models.
Although using image pre-trained models as the audio expert do not achieve the best results, they still show a reasonable performance.
Compared to the audio-only baseline using AST~\cite{gong21b_interspeech} as the audio expert in \tabref{cava_ablation:a}, using image-pre-trained as the audio expert shows favorable performance.
The results show that CAVA effectively utilizes the complementary knowledge of multiple experts through information exchange, leading to enhanced performance.

\subsubsection{Effect of time embedding}
To effectively employ positional embeddings across different modalities, we conduct an ablation study of embeddings strategies.
In \tabref{cava_ablation:e}, using conventional positional embeddings increases the learnable parameters by $18.8M$ compared to using no positional embedding. 
Despite the significant amount of additional parameters, the performance improvement is marginal.
When employing time embedding, we observe the highest top-1 accuracy of $60.3\%$ with $0.6M$ additional parameters only. 
The results suggest that using time embedding across different modalities provides clearer guidance for cross-attention between audio and visual experts.

\begin{table}[t]
\centering
\captionsetup{justification=justified, singlelinecheck=false}
\caption{\tb{Comparison with the state-of-the-arts on the EK100, SSV2 and K400.} We show the Top-1 accuracy on each dataset and the harmonic mean (H.M.) of the Top-1 accuracies. 
}

\resizebox{\linewidth}{!}{
\begin{tabular}{lcccccccccc}
\toprule
&&GFLOPs/& \multicolumn{3}{c}{EK100 Top-1} && \multicolumn{3}{c}{SSV2 \& K400 Top-1} & All \\
\cmidrule(lr){4-6} \cmidrule(lr){8-10} \cmidrule(lr){11-11} 
Method &&View& Verb & Noun & \cellcolor{gray!30}Act. && SSV2 & K400 & \cellcolor{gray!30}H.M. & \cellcolor{gray!30}H.M. \\
\midrule
CLIP*~\cite{radford2021learning} &&{140}& 54.9 & 52.7 & \cellcolor{gray!30}33.8 && 47.8 & 78.9 & \cellcolor{gray!30}59.5 & \cellcolor{gray!30}56.5 \\

EVL~\cite{lin2022frozen} &&{592}& - & - & \cellcolor{gray!30}- && 62.4 & 82.9 & \cellcolor{gray!30}71.2 & \cellcolor{gray!30}- \\

ST-Adapter~\cite{pan2022st} &&{607}& 67.6 & 55.0 & \cellcolor{gray!30}- && 69.5 & 82.7 & \cellcolor{gray!30}75.5 & \cellcolor{gray!30}67.3\\

AIM~\cite{yang2023aim} &&{404}& 64.8 & 55.5 & \cellcolor{gray!30}41.3* && 68.1 & 84.5 & \cellcolor{gray!30}75.4 & \cellcolor{gray!30}66.7 \\

\midrule
MBT~\cite{nagrani2021mbt} &&{936}& 64.8 & 58.0 & \cellcolor{gray!30}43.4 && - & 80.8 & \cellcolor{gray!30}- & \cellcolor{gray!30}- \\

ViViT FE~\cite{arnab2021vivit} &&{990}& 66.4 & 56.8 & \cellcolor{gray!30}44.0 && 65.9 & 81.7 & \cellcolor{gray!30}73.0 & \cellcolor{gray!30}66.6 \\

TimeSformer-B~\cite{bertasius2021space} &&{2380}& - & - & \cellcolor{gray!30}- && 62.4 & 80.7 & \cellcolor{gray!30}70.4 & \cellcolor{gray!30}- \\

MViT-B~\cite{fan2021multiscale} &&{170}& - & - & \cellcolor{gray!30}- && 67.7 & 80.2 & \cellcolor{gray!30}73.4 & \cellcolor{gray!30}- \\

MFormer-HR~\cite{patrick2021keeping} &&{1185}& 67.1 & 57.6 & \cellcolor{gray!30}44.1 && 68.1 & 80.2 & \cellcolor{gray!30}73.7 & \cellcolor{gray!30}67.3 \\

ORViT MF~\cite{herzig2022object} &&{-}& 68.4 & 58.7 & \cellcolor{gray!30}45.7 && 67.9 & - & \cellcolor{gray!30}- & \cellcolor{gray!30}- \\

Video Swin-L~\cite{liu2022video} &&{282}& - & - & \cellcolor{gray!30}- && 69.6 & 82.7 & \cellcolor{gray!30}75.8 & \cellcolor{gray!30}-\\

BEVT~\cite{wang2022bevt} &&{282}& - & - & \cellcolor{gray!30}- && 70.6 & 80.6 & \cellcolor{gray!30}75.3 & \cellcolor{gray!30}-\\

VideoMAE-B~\cite{tong2022videomae} &&{180}& 70.5 & 51.4 & \cellcolor{gray!30}41.7* && 70.8 & 81.5 & \cellcolor{gray!30}75.8 & \cellcolor{gray!30}66.6 \\

MeMViT~\cite{wu2022memvit} &&{59}& 70.6 & 58.5 & \cellcolor{gray!30}46.2 && - & - & \cellcolor{gray!30}- & \cellcolor{gray!30}-\\

OMNIVORE~\cite{girdhar2022omnivore} &&{-}& 69.5 & 61.7 & \cellcolor{gray!30}\best{49.9} && 71.4 & 84.0 & \cellcolor{gray!30}\second{77.2} & \cellcolor{gray!30}\second{70.8} \\

MTV-HR~\cite{yan2022multiview} &&{930}& 68.0 & 63.1 & \cellcolor{gray!30}48.6 && 68.5 & 82.4 & \cellcolor{gray!30}74.8 & \cellcolor{gray!30}69.8\\

\midrule
CAST &&{391}& 72.5 & 60.9 & \cellcolor{gray!30}\second{49.3} && 71.6 & 85.3 & \cellcolor{gray!30}\best{77.9} & \cellcolor{gray!30}\best{71.6} \\
\bottomrule
\end{tabular}
\vspace{\tabcapmargin}
\vspace{-1.5em}
}

\label{tab:main_rgb}
\raggedright
\vspace{0.5em}
\footnotesize{$^*$We conduct experiments with parameter-efficient tuning with adapters.}
\vspace{-0.5em}
\end{table}

\begin{table}[t]
\centering
\caption{\tb{Comparison with SOTA on EPIC-SOUNDS.} 
} 
\resizebox{0.8\linewidth}{!}{
\begin{tabular}{lcc}
\toprule
Method & Learnable Param. & Top-1 Acc. \\
\midrule
AST~\cite{gong21b_interspeech} & 87M & 54.9 \\
SSAST~\cite{gong2022ssast}     & 87M & 53.5 \\
Auditory SlowFast~\cite{kazakos2021slow} & 27M & 53.8 \\
MC3~\cite{chen2024soundingactions} & - & 56.0 \\
DiffSED~\cite{bhosale2024diffsed} & - & 56.9 \\
TIM(A+V)~\cite{chalk2024tim} & - & 58.3 \\
\midrule
Mirasol3B~\cite{piergiovanni2024mirasol3b}  & 3B & \best{78.2} \\
\midrule
CAST       & 45M & 47.8 \\
CAVA  & 44M & 60.3 \\
CA$^2$ST & 62M & \second{61.0} \\
\bottomrule
\end{tabular}
}
\label{tab:main_audio:epicsounds}
\vspace{-0.5em}
\end{table}

\begin{table}[t]
\centering
\caption{\tb{Comparison with SOTA on the UCF-101.} 
} 
\resizebox{0.8\linewidth}{!}{
\begin{tabular}{lcc}
\toprule
Method & Learnable Param.& Top-1 Acc. \\
\midrule
AVSlowFast~\cite{xiao2020audiovisual} & 39M & 87.0 \\
XDC~\cite{alwassel2020self} & 65M & 95.5 \\
MMV~\cite{alayrac2020self} & 94M & 95.2 \\
AVID~\cite{morgado2021audio} & 47M & 91.5 \\
GDT~\cite{patrick2021compositions} & 33M & 95.2 \\
\sred{Noise-tolerant~\cite{han2024noise}} & - & 87.3 \\
XKD~\cite{sarkar2024xkd} & 87M  & 94.1 \\
MoMA~\cite{wang2024moma} & 17M & 92.7 \\
\midrule
CAST  & 45M & \second{96.9} \\
CAVA  & 44M & 96.1 \\
CA$^2$ST & 62M & \best{97.2} \\
\bottomrule
\end{tabular}
}
\label{tab:main_audio:ucf}
\end{table}

\begin{table}[t]
\centering
\caption{\tb{Comparison with SOTA on the VGG-Sound.} 
} 
\resizebox{0.8\linewidth}{!}{
\begin{tabular}{lcc}
\toprule
Method & Learnable Param. & Top-1 Acc. \\
\midrule
MBT~\cite{nagrani2021mbt}   & - & 64.1 \\
MMT(A+V)~\cite{zhu2024efficient}   & 52M & 66.2 \\
MAViL(A+V)~\cite{huang2024mavil}  & 86M & 67.1 \\
EquiAV(A+V)~\cite{pmlr-v235-kim24v}  & - & 67.1 \\
CAV-MAE~\cite{gong2023contrastive} & 164M & 65.5 \\
AudiovisualMAE~\cite{georgescu2023audiovisual} & 36M & 65.0 \\
CrossMAE~\cite{guo2024crossmae}  & - & 67.0 \\
\midrule
Mirasol3B~\cite{piergiovanni2024mirasol3b}  & 3B &  \best{69.8} \\
\midrule
CAST & 45M & 54.7 \\
CAVA  & 44M & 68.2 \\
CA$^2$ST & 62M & \second{68.3} \\
\bottomrule
\end{tabular}
\label{tab:main_audio:vgg}
}
\vspace{-0.5em}
\end{table}

\begin{table}[t]
\centering
\caption{\tb{Comparison with SOTA on KineticsSound.} 
} 
\resizebox{0.75\linewidth}{!}{
\begin{tabular}{lcc}
\toprule
Method & Learnable Param. & Top-1 Acc. \\
\midrule
\sred{Noise-tolerant~\cite{han2024noise}} & - & 84.6 \\
MoMA~\cite{wang2024moma} & 17M & 91.1 \\
MBT~\cite{nagrani2021mbt} & - & 85.0 \\
MMT(A+V)~\cite{zhu2024efficient}  & 52M & 92.3 \\
\midrule
Mirasol3B~\cite{piergiovanni2024mirasol3b}  & 3B & 90.1 \\
\midrule
CAST & 45M & 91.6 \\
CAVA & 44M & \second{92.9} \\
CA$^2$ST & 62M & \best{93.3} \\
\bottomrule
\end{tabular}
\label{tab:main_audio:kineticsounds}
\vspace{-1.5em}
}
\end{table}

\vspace{\secmargin}
\subsection{Comparison with action recognition state-of-the-art} 
\label{sec:comparison}

In this section, we evaluate the performance of CAST and state-of-the-art methods in terms of balanced spatio-temporal understanding on multiple datasets, as shown in \tabref{main_rgb}. For each method, we report the top-1 accuracy of each task and the harmonic mean of top-1 accuracies for i) SSV2~\cite{goyal2017something}, and K400~\cite{kay2017kinetics}, and ii ) EK100~\cite{damen2022rescaling} verb, EK100 noun, SSV2, and K400. 
For the comparison with state-of-the-art models, we tune the hyperparameters \eg{learning rate} to optimize performance. 
For fair comparisons of the computation complexity, we show the GFLOPs/View. In cases where a compared method shows various GFLOPs/View depending on the dataset, we specifically note the lowest GFLOPs/View value for reference. 
We observe that the CLIP\cite{radford2021learning}-based methods (the first group in Table \ref{tab:main_rgb}), AIM~\cite{yang2023aim} achieves favorable performance on the static-biased K400 dataset, with $84.5\%$ accuracy.
However, AIM shows a relatively lower performance of $68.1\%$ on the temporal-biased SSV2 dataset. 
On the other hand, VideoMAE~\cite{tong2022videomae} shows $70.8\%$ accuracy on the SSV2 dataset, which is more competitive than AIM. 
However, VideoMAE shows a lower accuracy of $81.5\%$ on the K400 dataset, less competitive than AIM.
The results indicate that these models tend to have a imbalanced spatio-temporal understanding.

Our proposed method, CAST, demonstrates favorable performance on both the SSV2 ($71.6\%$) and K400 ($85.3\%$) datasets, resulting in a harmonic mean of $77.9\%$, which is higher than that of AIM ($75.4\%$) and VideoMAE ($75.8\%$). CAST shows a more balanced spatio-temporal understanding than the existing methods. Additionally, CAST shows favorable performance in fine-grained action recognition on the EK100 dataset except for OMNIVORE~\cite{girdhar2022omnivore}.
It is worth noting that OMNIVORE learns image, video, and depth representations jointly, making its training process more complex.

In terms of the overall harmonic mean of EK100 verb, EK100 noun, SSV2, and K400 accuracies, CAST shows the best performance of $71.6\%$. 
The results highlight the effectiveness of CAST. By exchanging information between spatial and temporal experts, our CAST shows a favorable balanced spatio-temporal understanding performance.

\vspace{\secmargin}
\subsection{Comparison with audio-visual recognition state-of-the-art} 
\label{sec:cava_comparison}
In this section, we compare the performance of CAVA and \caast{} with state-of-the-art methods across multiple datasets, as shown in \tabref{main_audio:epicsounds}-\tabref{main_audio:kineticsounds}. 
We report the top-1 accuracy the number of learnable parameters on four audio-visual recognition datasets: i) EPIC-SOUNDS~\cite{huh2023epic}, ii) UCF-101~\cite{soomro2012dataset}, iii)VGG-Sound~\cite{chen2020vggsound}, and iv) KineticsSound~\cite{xiao2020audiovisual}.
We choose these datasets to demonstrate the robustness and versatility of our architecture in integrating audio-visual modalities for holistic video action recognition.
CAVA demonstrates robust performance across all datasets, consistently shows favorable performance compared to state-of-the-art methods. 

As shown in \tabref{main_audio:epicsounds}, CAVA achieves a top-1 accuracy of $60.3\%$, surpassing existing methods such as AST~\cite{gong21b_interspeech}, SSAST~\cite{gong2022ssast} and Auditory SlowFast~\cite{kazakos2021slow} on the EPIC-SOUNDS dataset.
CAVA significantly outperforms AST, our audio expert, showing the effectiveness of information exchange in improving audio-visual understanding.
Specifically, unlike the other datasets, EPIC-SOUNDS consists of egocentric videos where temporal information plays a crucial role. 
As a result, \caast{}, which includes a temporal expert in addition to CAVA, achieves the second best performance on this dataset.

On the UCF-101 dataset~\cite{soomro2012dataset}, \caast{} achieves the top-1 accuracy of $97.2\%$ as shown in \tabref{main_audio:ucf}, outperforming all compared methods. CAST shows the second-best performance of $96.9\%$, followed closely by CAVA at $96.1\%$. 
Compared to XDC~\cite{alwassel2020self} and MMV~\cite{alayrac2020self}, CAVA achieves higher performance with significantly fewer learnable parameters at 44M (\vs 65M of XDC, 94M of MMV). 
Since UCF-101 has a relatively higher emphasis on visual information, CAST designed with strong spatial and temporal experts shows a strong performance. 
The results indicate that audio cue is useful for video recognition, with \caast{} benefiting from the integration of spatial, temporal, and audio experts for a more holistic understanding.

Although Mirasol3B outperforms our CAVA on the EPIC-SOUNDS and VGG-Sound~\cite{chen2020vggsound} datasets, 
it is important to note that Mirasol3B~\cite{piergiovanni2024mirasol3b} is a large foundation model pretrained on 3M video-text pairs (VTP~\cite{alayrac2022flamingo} while CAVA is not pretrained on any video-text pairs), with a lot of parameters (3B \vs 44M$\sim$62M), 
using a larger number of input frames (128 frames \vs 16),
and higher input resolution ($448 \times 448$ \vs $224 \times 224$).
Considering these substantial disadvantages in the number of parameters, amount of pertaining data, and input size, \caast{} demonstrates competitive performance on the VGG-Sound~\cite{chen2020vggsound} following Mirasol by only 1.5 points, with CAVA following closely in third as shown in \tabref{main_audio:vgg}.
Remarkably, as shown in \tabref{main_audio:kineticsounds}, \caast{} surpasses Mirasol3B on the KineticsSound~\cite{xiao2020audiovisual} dataset by 3.2 points, achieving the best accuracy ($93.3\%$). 

Overall, the consistent favorable performances across multiple datasets underscore the effectiveness of the proposed method.
The results showcase the capability of CAVA and \caast{} for holistic audio-visual understanding.

\subsection{{Comparison of model efficiency and complexity}}
\label{sec:complexity}
\begin{table}[t]
    \centering
    \captionsetup{justification=justified}
    \caption{\textbf{Comparison of model complexity and accuracy on EPIC-SOUNDS.}}
    \resizebox{0.9\linewidth}{!}{
    \begin{tabular}{l ccccc}
    \toprule
    Method & Param. &Throughput (V/s)&	Latency (ms)& GFLOPs & Acc.\\
    \midrule
    VideoMAE-B & 87M& 67&17.6&192& 31.8  \\
    CLIP-B &86M& 81&14.7& 152& 31.9 \\
    \midrule
    CAST& 217M&37&38.5& 391& 47.8\\
    CAVA& 216M&65&36.6& 236&60.3 \\
    CA$^2$ST& 320M&30&54.8& 454&61.0 \\
    \bottomrule
    \end{tabular}
    }
    \label{tab:complexity}
    \vspace{-0.5em}
\end{table}
\begin{table}[t]
    \centering
    \caption{\textbf{Comparison with state-of-the-art methods on the EK100, HD-EPIC, and EK100 Cross-Subject split.} 
}
    \resizebox{0.9\linewidth}{!}{
    \begin{tabular}{l  c ccc ccc c}
    \toprule
    & \multicolumn{3}{c}{EK100 Top-1 Acc.} & \multicolumn{3}{c}{HD-EPIC Top-1 Acc.} & Cross-Subject\\
    \cmidrule(lr) {2-4}
    \cmidrule(lr) {5-7}
    Method  & Verb & Noun & Act. & Verb & Noun & Act. &  Act.\\
    \midrule
    Omnivore~\cite{girdhar2022omnivore}&69.5 &61.7&\second{49.9}&19.5 & 17.1&8.7 &  28.7\\
    MotionFormer-HR~\cite{patrick2021keeping}  &67.0&58.5&44.5& 35.7&20.0&10.2 &  32.2\\
    VideoMAE-L~\cite{tong2022videomae}  & -&-&-&47.5 & 29.4 &17.9 & 29.3 \\
    TIM~\cite{chalk2024tim} &77.1&67.2&\best{57.5}& 51.3 & 36.1 &\best{23.4} & \textbf{44.6} \\
    \midrule
    CAST & 72.5&60.9&49.3&47.1 &30.3&\second{18.4}& 39.2 \\
    CAVA & 66.5&54.1& 41.5 &35.7&22.3&10.6& 35.1 \\
    C$\text{A}^2$ST & 72.9&60.7&49.6& 46.7 & 28.8&17.4& \second{42.4} \\
    \bottomrule
    \end{tabular}
    }
    \label{tab:hd-epic-only}
    \vspace{-0.5em}
\end{table}

We evaluate the computational cost and efficiency of our models—CAST, CAVA, and CA$^2$ST—on the EPIC-SOUNDS~\cite{huh2023epic} dataset, with results summarized in \tabref{complexity}. 
We report standard metrics including total parameters, throughput (videos/sec), latency, GFLOPs, and Top-1 accuracy. 
For comparison, we also include adapter-tuned single-stream models, VideoMAE-B~\cite{tong2022videomae} and CLIP-B~\cite{radford2021learning}. 
While these baselines are lightweight and fast (14.7–17.6ms latency), they underperform (31.8–31.9\% accuracy) due to limited modality usage.
Our models offer a flexible trade-off between performance and efficiency by leveraging modality-specific experts.
CAST (visual-only) achieves strong performance (47.8\%) with moderate latency and compute.
CAVA (audio + visual) balances efficiency and accuracy (60.3\%, 36.6ms latency, 65 V/s throughput), making it ideal for resource-constrained scenarios.
\caast{} (audio + spatial + temporal) delivers the best accuracy (61.0\%) at higher cost (54.8ms latency), suitable when performance is crucial.
This modular design enables tailored deployment based on application requirements.

\subsection{{Analyzing cross-modal integration by B-CA module}}
\label{sec:analysis_bca}

We analyze the B-CA module to understand how it aligns semantics between experts, focusing on information exchange between the audio expert (AST~\cite{gong21b_interspeech}) and the visual expert (CLIP~\cite{radford2021learning}). 
This is motivated by the relatively limited information in audio compared to rich spatio-temporal visual features. 
We conduct both quantitative (layer-wise cross-attention entropy) and qualitative (attention map visualization) analyses using the VGG-Sound~\cite{chen2020vggsound} dataset.

\subsubsection{Quantitative analysis via layer-wise attention entropy}
\label{sec:entropy}
\begin{figure}[t]
    \centering
    \includegraphics[width=0.8\linewidth]{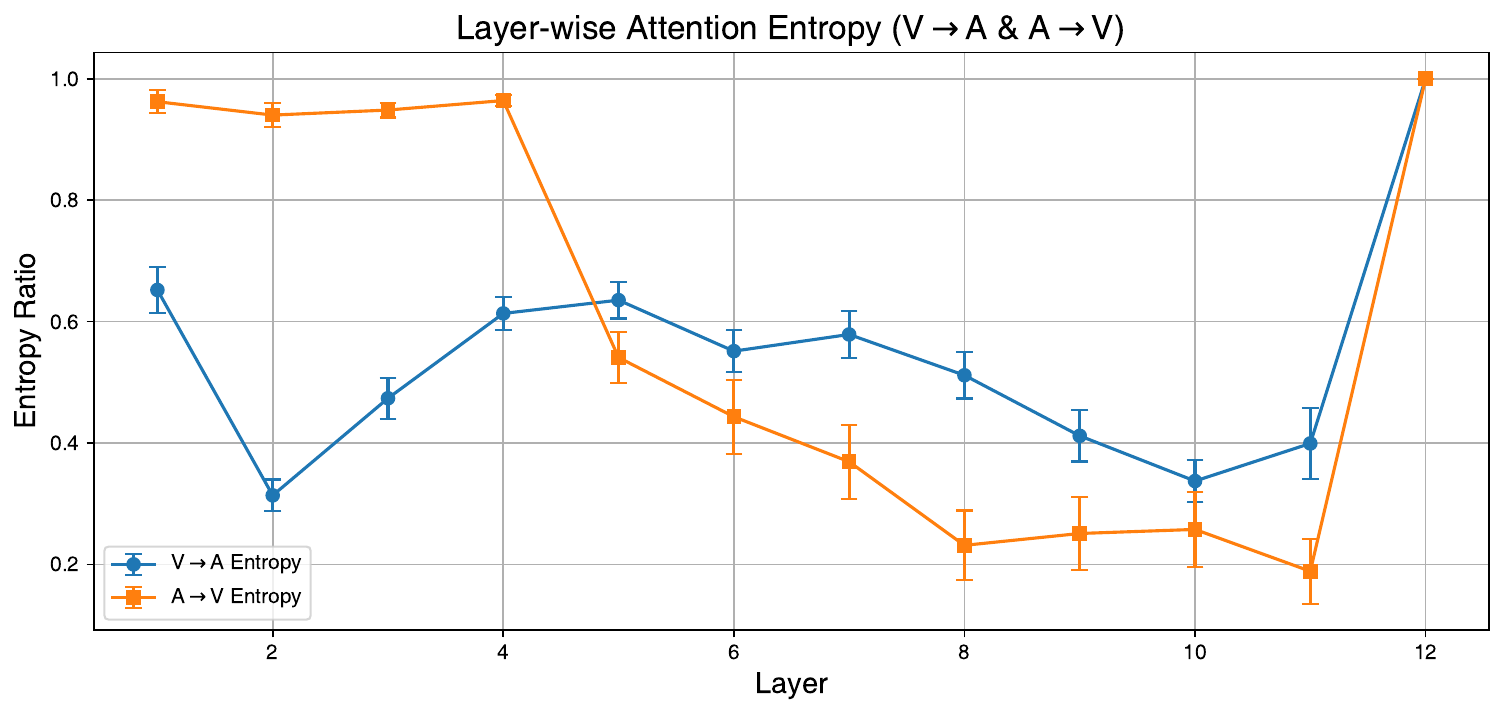}
    \figcaption{Layer-wise Cross-Attention Entropy on VGG-Sound}{
    We plot the entropy ratio of Spatial-to-Audio (S2A) and Audio-to-Spatial (A2S) cross attention weights in CAVA across layers, where values closer to 1 indicate uniform attention and lower values reflect more focused attention. 
    }
    \label{fig:layerwise_entropy_f1_vggsound}
\end{figure}

To analyze how the B-CA module aligns semantics across experts, we visualize cross-attention entropy across layers. 
As shown in \figref{layerwise_entropy_f1_vggsound}, entropy drops notably at layers 6 and 9 in both directions (audio↔visual), indicating more focused cross-modal attention at these points. 
The result suggests the model adaptively fuses semantics based on modality-specific information. 
These findings offer insight into how B-CA balances and aligns diverse modality features.

\begin{figure*}[!t]
    \centering
    \includegraphics[width=0.8\linewidth]{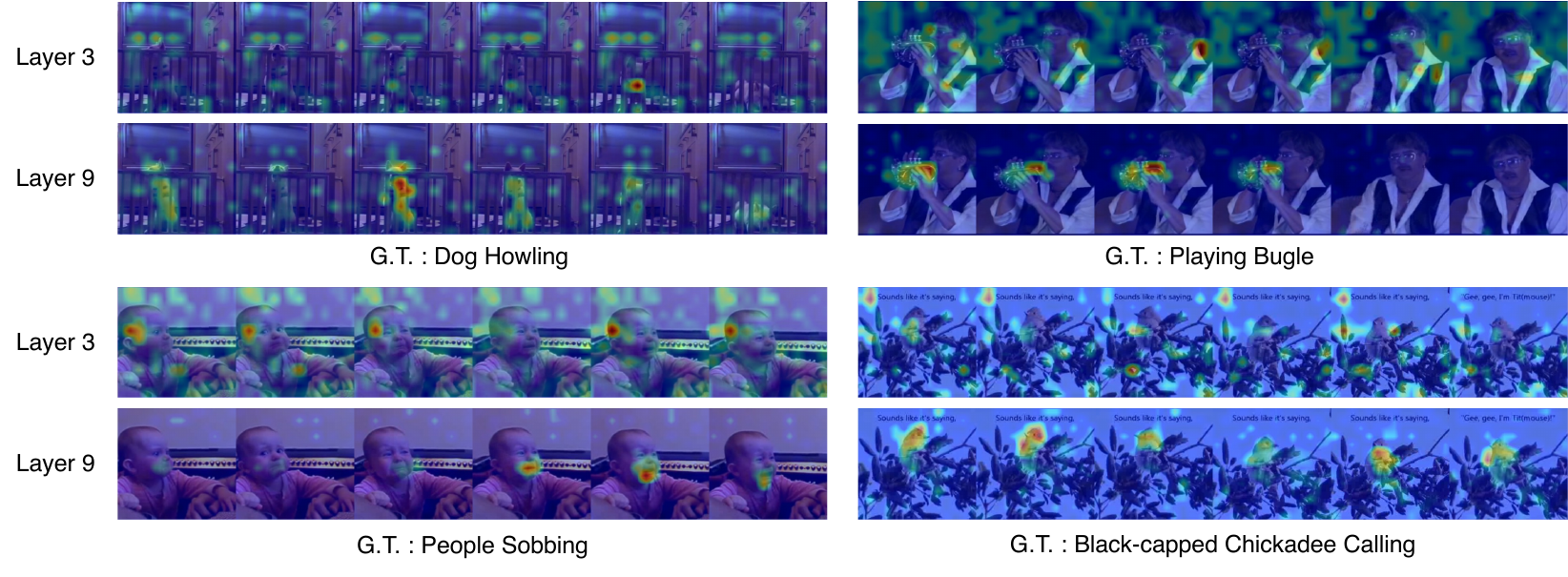}
    \vspace{-1em}
    \figcaption{Visualization of Spatial-to-Audio (S2A) attention map}{We visualize S2A attention maps from CAVA for \emph{Dog Howling}, \emph{Playing Bugle}, \sred{\emph{People Sobbing}, and \emph{Black-capped Chickadee Calling}.}
    While layer 3 shows broadly distributed attention, layer 9 focuses on semantically meaningful regions, such as the sound source or interacting human, highlighting B-CA's ability to transfer knowledge between experts.
    }
    \label{fig:v2s_attn_map}
    \vspace{-1em}
\end{figure*}
\subsubsection{Qualitative analysis via cross-attention map visualization}
\label{sec:attnmap}
To understand how the B-CA module aligns audio and visual information, we analyze attention maps from the Spatial-to-Audio (S2A) cross-attention layers in CAVA, focusing on how the audio expert incorporates visual cues—especially important given audio’s lower information density. 
Based on entropy patterns in \figref{layerwise_entropy_f1_vggsound}, we visualize layers 3 (high entropy) and 9 (low entropy) in the visual-to-audio direction.

\figref{v2s_attn_map} shows that early layers exhibit diffuse attention, while deeper layers focus on semantically relevant regions. 
For example, in \emph{Dog Howling}, layer 9 attention tracks the moving subject; in \emph{Play the bugle}, it highlights both the bugle and the person. 
Similarly, in \emph{People Sobbing}, attention peaks as crying begins; in \emph{Black-capped Chickadee Calling}, it remains focused on the bird even against dense foliage.
The results suggest B-CA enables the audio expert to selectively use informative visual cues, improving recognition in multi-modal settings.

\subsection{{Robustness Evaluation}}
\label{sec:robustness}
\subsubsection{Action recognition in unseen scenario}
We assess the robustness and generalization of our method in two challenging unseen scenarios: i) EK100$\rightarrow$HD-EPIC and ii) EK100 cross-subject.
In the EK100$\rightarrow$HD-EPIC setting, we train a model on the EK100~\cite{damen2022rescaling} and test it on the HD-EPIC~\cite{hd-epic}, a domain-shifted dataset. 
In the cross-subject setting, we test on participants not seen during training.
As shown in \tabref{hd-epic-only}, CAST and CA$^2$ST generalize well in both settings. CAST ranks second on HD-EPIC, outperforming Omnivore~\cite{girdhar2022omnivore} by 9.7 points. CA$^2$ST achieves the second-best result in the cross-subject setting, showing strong subject-level generalization.
While TIM~\cite{chalk2024tim} slightly outperforms our models, it uses multiple large-scale transformer backbones. 
In contrast, our models are more efficient while maintaining competitive performance.
In summary, the results demonstrate the proposed methods' strong generalization under distribution shifts.

\subsubsection{Evaluating robustness on long video sequences}
\begin{table}[t]
    \centering
    \caption{\textbf{Comparison with state-of-the-art methods on the ActivityNet dataset.} 
    }
    \resizebox{0.9  \linewidth}{!}{
    \begin{tabular}{l cc c c}
    \toprule
    Method & TFLOPs&Backbone &\#F &  Top-1 Acc.\\ 
    \midrule
    DSN~\cite{dsn} &-&R(2+1)D-34&80& 82.6\\
    SMART~\cite{gowda2021smart}&- &-&24& 84.4\\
    MARL\cite{marl} &-&SEResNeXt-152&120& 85.7\\
    TSQNet\cite{tsqnet} &-&Swin-L&50& 88.7\\
    NSNet\cite{xia2022nsnet}&-&Swin-L &100& 90.2\\
    InternVideo2~\cite{wang2024internvideo2}&36.3 &InternVL-6B~\cite{chen2024internvl}&16& \best{95.9} \\
    \midrule
    CAST& 1.2 &VideoMAE-B\&CLIP-B&48& \second{91.3} \\
    \bottomrule
    \end{tabular}
    }
    \label{tab:act_only}
    \vspace{-0.5em}
\end{table}
\begin{table}[t]
    \centering
    \caption{\textbf{Comparison with state-of-the-art methods on EPIC-SOUNDS and HD-EPIC-SOUNDS datasets.} 
    }
    \resizebox{0.9\linewidth}{!}{
    \begin{tabular}{l c c c c c}
    \toprule
    & & Total &  & EPIC-SOUNDS&HD-EPIC-SOUNDS \\
    Method  & Modality & Param. & GFLOPs  & Top-1 Acc. & Top-1 Acc.\\
    \midrule
    SSAST~\cite{gong2022ssast} & A & - & - & 53.5 & 25.1 \\
    TIM~\cite{chalk2024tim} & A & 64M & 116 & 55.7 & 26.9 \\
    Auditory SlowFast~\cite{kazakos2021slow} & A & 26M & 4 &  53.8 & 27.9 \\
    TIM~\cite{chalk2024tim} & A+V & 435M & 46017*  & 58.3 & \textbf{31.9} \\
    \midrule
    CAST & V & 217M & 391 &  47.8 & 27.8 \\
    CAVA & A+V & 216M & 236 &  \second{60.3} & \second{28.6} \\
    C$\text{A}^2$ST & A+V & 320M & 454 & \textbf{61.0} & 28.1 \\
    \bottomrule
    \end{tabular}
    }
    \label{tab:hd-epic-sounds}
    \raggedright
    
    \scriptsize{* TIM evaluates 50 clips / modality from a 30s untrimmed window (0.6s stride).}
    \vspace{-0.5em}
\end{table}

To evaluate robustness on long videos, we evaluate CAST on ActivityNet~\cite{caba2015activitynet}, where the average video length is 65 seconds—much longer than K400’s 10-second clips.
We use 48 input frames to capture extended temporal context.
As shown in \tabref{act_only}, CAST achieves 91.3\% Top-1 accuracy, outperforming temporal sampling-based methods like TSQNet~\cite{tsqnet} (88.7\%) and NSNet~\cite{xia2022nsnet} (90.2\%), despite using fewer frames. 
Although InternVideo2~\cite{wang2024internvideo2} outperforms CAST, it relies on a massive 6B-scale backbone and incurs 36.27 TFLOPs— $\sim30\times$ higher than CAST.
These results demonstrate that CAST generalizes well to long-video scenarios and supports efficient temporal modeling beyond short clips. 
While not explicitly designed for long-range reasoning, CAST’s strong performance suggests potential for future applications in tasks like memory retrieval and video QA.

\subsubsection{Audio-Visual recognition in unseen scenario}
We evaluate our method’s generalization by training on EPIC-SOUNDS~\cite{huh2023epic} and testing on both EPIC-SOUNDS (in-domain) and HD-EPIC-SOUNDS~\cite{hd-epic}, which introduces domain shifts through newly collected videos in unseen environments.
\tabref{hd-epic-sounds} compares our models with state-of-the-art methods. 
All methods are trained on EPIC-SOUNDS and evaluated on HD-EPIC-SOUNDS without fine-tuning for fair comparison. 
We also report GFLOPs and input modalities for efficiency analysis.

As discussed in \secref{cava_comparison}, both CAVA and CA$^2$ST perform favorably on EPIC-SOUNDS and remain competitive under domain shifts, demonstrating robustness.
Notably, the visual modality becomes more important on HD-EPIC-SOUNDS. CAST underperforms audio-only models on EPIC-SOUNDS but surpasses them on HD-EPIC-SOUNDS. 
For instance, TIM(A) outperforms CAST by 7.9 points on EPIC-SOUNDS but falls behind by 0.9 points on HD-EPIC-SOUNDS. 
Adding a large-scale visual encoder to TIM(A) improves accuracy by 2.6 points on EPIC-SOUNDS and 5.0 points on HD-EPIC-SOUNDS, suggesting greater reliance on visual cues in out-of-domain settings.
While TIM(A+V) achieves the highest accuracy (31.9\%) on HD-EPIC-SOUNDS, it requires over 46TFLOPs. 
In contrast, CAVA is the second-best (28.6\%) with only 236 GFLOPs, highlighting the efficiency and robustness of our approach under domain shifts.

\subsubsection{Audio-Visual recognition under audio corruption}
\begin{table}[t]
    \centering
    \captionsetup{justification=justified, singlelinecheck=false}
    \captionsetup{justification=justified}
    \caption{\textbf{Robustness evaluation under diverse audio-visual corruptions on EPIC-SOUNDS.}
    }
    \resizebox{0.6\linewidth}{!}{
    \begin{tabular}{lcc}
    \toprule
    Method &Noise Type& Top-1 Acc. \\
    \midrule
    AST~\cite{gong21b_interspeech} & -& 54.9 \\
    MC3~\cite{chen2024soundingactions} & - & 56.0 \\
    DiffSED~\cite{bhosale2024diffsed}  & - & 56.9 \\
    TIM(A+V)~\cite{chalk2024tim}   & -& 58.3 \\
    CAVA & -& 60.3  \\
    \midrule
    CAVA  &Misalignment & 59.7 \\
    CAVA   &Dropout& 58.1 \\
    CAVA  &Gaussian noise & 58.2\\
    CAVA  &Pink noise  &58.4 \\
    \bottomrule
    \end{tabular}
    }
    \label{tab:audio-noise}
    \vspace{-1.2em}
\end{table}
We evaluate the robustness of CAVA under realistic audio corruption scenarios, including misalignment, dropout, and noise. 
Specifically, we evaluate four types of corruption scenarios on the EPIC-SOUNDS dataset:
(i) Misalignment (random audio shift of –2 to +2 seconds),
(ii) Dropout (20\% random audio removal),
(iii) Gaussian noise, and
(iv) Pink noise (natural background noise).
As shown in \tabref{audio-noise}, CAVA maintains strong performance under all scenarios, with accuracy drops of only 0.6–1.9 points. 
Even under the most severe case (Dropout), it achieves 58.1\% accuracy—comparable to TIM(A+V)~\cite{chalk2024tim} at 58.3\%.
These results demonstrate CAVA’s robustness to audio degradation and the effectiveness of its audio-visual integration.
\vspace{-1em}
 
\section{Conclusions}
\label{sec:conclusions}

We tackle the challenge of balanced spatio-temporal understanding in action recognition by proposing CAST, which combines spatial and temporal experts with cross-attention for effective information exchange. 
Experiments across diverse datasets show that CAST outperforms individual experts and existing methods in balanced spatio-temporal performance.
To handle more complex scenarios, we extend CAST to multi-modal frameworks: CAVA, which integrates an audio expert, and CA$^2$ST, which combines audio, spatial, and temporal experts for holistic video understanding. 
Evaluations on EPIC-SOUNDS, UCF-101, VGG-Sound, KineticsSound, and HD-EPIC-SOUNDS confirm their strong audio-visual performance.
Ablation and qualitative analyses highlight the B‑CA module’s ability to efficiently align multi-modal cues. 
Overall, our method shows strong potential for robust and comprehensive video understanding.

\section*{Acknowledgments}
This work was supported by a grant from Kyung Hee University in 2024. (KHU-20241094)

\bibliographystyle{IEEEtran}
\bibliography{TPAMI_main}



 

\vspace{-3em}
\begin{IEEEbiography}[{\includegraphics[width=1in,height=1.25in,clip,keepaspectratio]{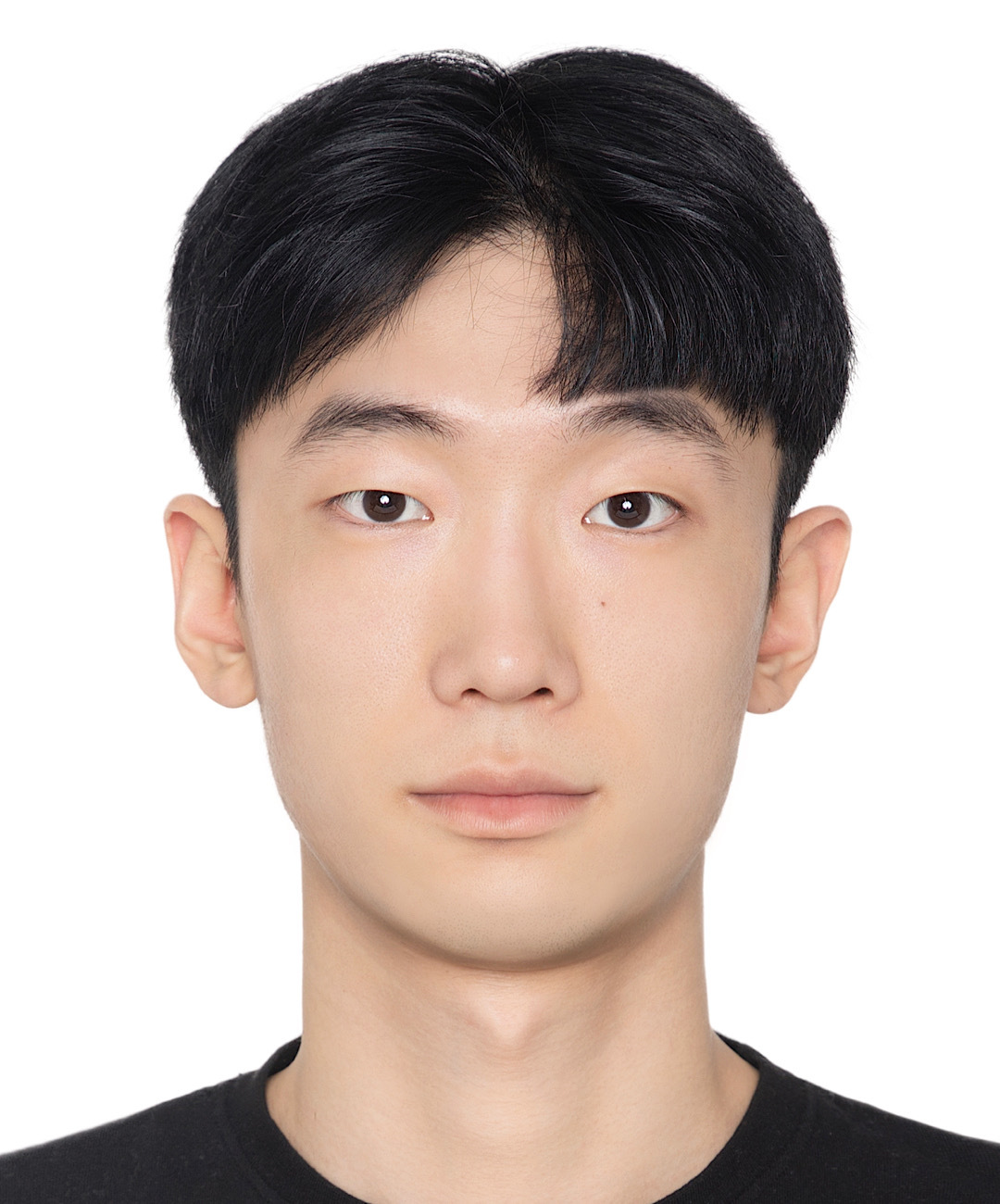}}]{Jongseo Lee}
received the B.S. degree in Biomedical Engineering and Electronic Engineering and the M.S. degree in Computer Science from Kyung Hee University, Republic of Korea, in 2023 and 2025, respectively. He is currently a post-master researcher at the Vision and Learning Lab, Kyung Hee University. His research interests include video understanding, multi-modal learning, and explainable AI.
\end{IEEEbiography}
\vspace{-2em}
\begin{IEEEbiography}[{\includegraphics[width=1in,height=1.25in,clip,keepaspectratio]{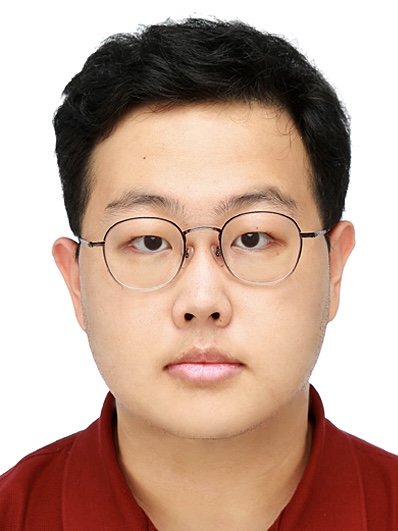}}]{Joohyun Chang}
 received the B.S. degree in Computer Science and Engineering from KyungHee University in 2025. He is currently pursuing an M.S. degree in School of Computing at the Korea Advanced Institute of Science and Technology (KAIST), where his research focuses on video understanding, multi-modal learning and Visual Language Models.
\end{IEEEbiography}
\vspace{-2em}

\begin{IEEEbiography}[{\includegraphics[width=1in,height=1.25in,clip,keepaspectratio]{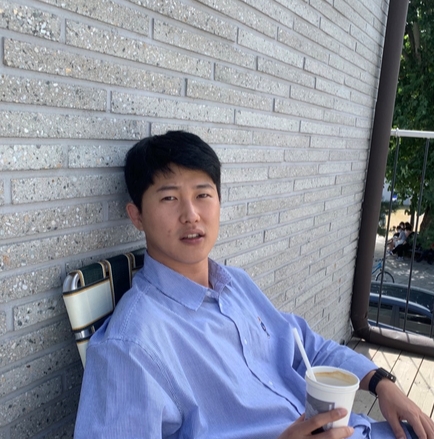}}]{Dongho Lee}
 received M.S. degree from the Department of Computer Science and Engineering, Kyung Hee University in 2023. He is currently a researcher at CJ AI Center. His research focuses on video understanding, foundation models, and multi-modal learning.
\end{IEEEbiography}
\vspace{-2em}

\begin{IEEEbiography}[{\includegraphics[width=1in,height=1.25in,clip,keepaspectratio]{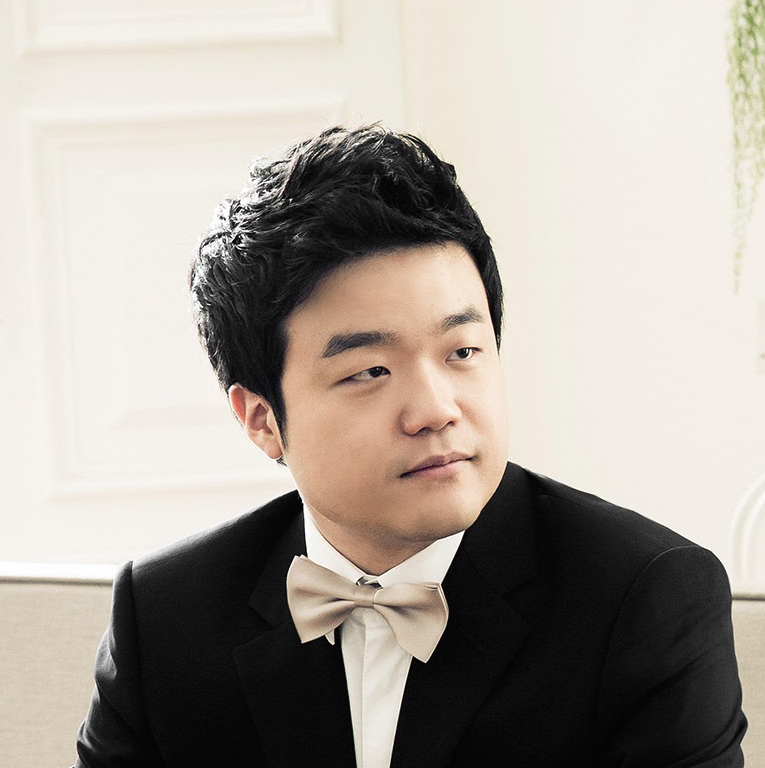}}]{Jinwoo Choi}
is an Assistant Professor of Computer Science and Engineering, Kyung Hee University, Republic of Korea. He obtained a Ph.D. from the Bradley Department of Electrical and Computer Engineering, Virginia Tech, in 2020. He was a computer vision researcher at Electronics and Telecommunications Research Institute (ETRI), Republic of Korea from 2010 to 2015. His research interests include video representation learning, action recognition, domain adaptation, debiasing, multi-modal learning, scene understanding, and explainable AI.
\end{IEEEbiography}

\vfill




\end{document}